%% file: main.tex
\documentclass[10pt,twocolumn,letterpaper]{article}

\usepackage[pagenumbers]{cvpr} 
\usepackage[table]{xcolor}
\usepackage{multirow,makecell,rotating}
\usepackage{algorithm}
\usepackage[noend]{algpseudocode}

\makeatletter
\renewcommand{\paragraph}{%
  \@startsection{paragraph}{4}%
  {\z@}{0.2ex \@plus 1ex \@minus .2ex}{-1em}%
  {\normalfont\normalsize\bfseries}%
}

\usepackage{booktabs,subcaption,amsfonts,dcolumn}
\usepackage{comment}
\usepackage{times}
\usepackage{epsfig}
\usepackage{amsmath}
\usepackage{amssymb}
\usepackage{algorithm}
\usepackage{algpseudocode}
\usepackage{booktabs}
\usepackage{subcaption}
\usepackage{graphicx}
\usepackage{enumitem}
\usepackage{scrextend}

\makeatletter
\renewcommand{\paragraph}{%
  \@startsection{paragraph}{4}%
  {\z@}{0.2ex \@plus 1ex \@minus .2ex}{-0.75em}%
  {\normalfont\normalsize\bfseries}%
}

\definecolor{cvprblue}{rgb}{0.21,0.49,0.74}
\usepackage[pagebackref,breaklinks,colorlinks,citecolor=cvprblue]{hyperref}

\def\methodName{RAVE}

\title{RAVE: \underline{Ra}ndomized Noise Shuffling for Fast and Consistent \underline{V}ideo \underline{E}diting with Diffusion Models}

\author{Ozgur Kara$^1$\thanks{}\qquad Bariscan Kurtkaya$^2$\footnotemark[1] \footnotemark[2]\qquad
Hidir Yesiltepe$^4$\qquad
James M. Rehg$^{1,3}$\qquad Pinar Yanardag$^4$
\\\\
{$^1$Georgia Tech}
{$^2$KUIS AI Center}
{$^3$UIUC}
{$^4$Virginia Tech}
\\
{\small \texttt{okara7@gatech.edu, bkurtkaya23@ku.edu.tr, hidir@vt.edu, jrehg@uiuc.edu, pinary@vt.edu}} \qquad
\\\\
Project Webpage: \href{https://rave-video.github.io/}{https://rave-video.github.io}}

\begin{document}

\twocolumn[{%

\renewcommand\twocolumn[1][]{#1}%
\maketitle
\vspace{-2.5em}
\begin{center}
    \centering
    \captionsetup{type=figure}
    \includegraphics[width=1.0\textwidth]{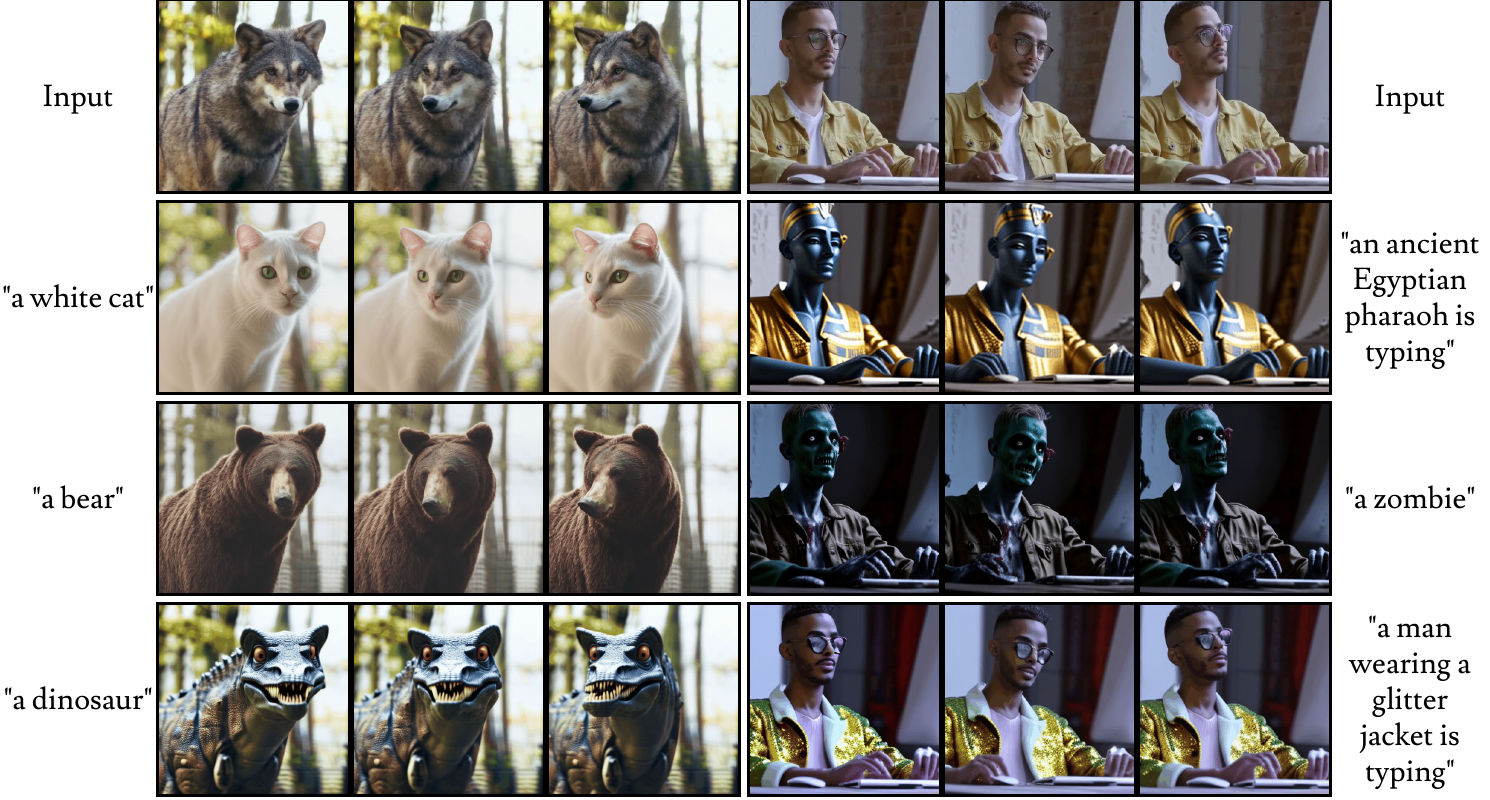}
    \captionof{figure}{\textbf{\methodName} is a lightweight and fast video editing method that enhances temporal consistency in video edits, utilizing pre-trained text-to-image diffusion models. It is capable of modifying local attributes, like changing a person's \textit{jacket} (bottom right), and can also handle complex shape transformations, such as turning a \textit{wolf} into a \textit{dinosaur} (bottom left).}
    \label{fig:teaser}
\end{center}%
}]

\renewcommand{\thefootnote}{\fnsymbol{footnote}}
\footnotetext[1]{~Joint co-author.}
\footnotetext[2]{~B. Kurtkaya worked on this project as an intern at Virginia Tech.}

\input{sec_arxiv/0_abstract}    
\input{sec_arxiv/1_intro}

\input{sec_arxiv/2_related}

\input{sec_arxiv/3_methodology}

\input{sec_arxiv/4_dataset}

\input{sec_arxiv/5_exp}

\input{sec_arxiv/6_discussion}

{
    \small
    \bibliographystyle{ieeenat_fullname}
    \bibliography{main}
}

\input{sec_arxiv/supp}

\end{document}

%% file: sec_arxiv/0_abstract.tex
\begin{abstract}

Recent advancements in diffusion-based models have demonstrated significant success in generating images from text. However, video editing models have not yet reached the same level of visual quality and user control. To address this, we introduce RAVE, a zero-shot video editing method that leverages pre-trained text-to-image diffusion models without additional training. RAVE takes an input video and a text prompt to produce high-quality videos while preserving the original motion and semantic structure.  
It employs a novel noise shuffling strategy, leveraging spatio-temporal interactions between frames, to produce temporally consistent videos faster than existing methods. It is also efficient in terms of memory requirements, allowing it to handle longer videos.  RAVE is capable of a wide range of edits, from local attribute modifications to shape transformations. In order to demonstrate the versatility of RAVE, we create a comprehensive video evaluation dataset ranging from object-focused scenes to complex human activities like dancing and typing, and dynamic scenes featuring swimming fish and boats. Our qualitative and quantitative experiments highlight the effectiveness of RAVE in diverse video editing scenarios compared to existing methods. Our code, dataset and videos can be found in \href{https://rave-video.github.io/}{https://rave-video.github.io}.
\end{abstract}

%% file: sec_arxiv/1_intro.tex
\section{Introduction}
\label{sec:intro}

Diffusion-based generative models \cite{sohl2015deep,ho2020denoising} demonstrate outstanding capabilities in generating and editing diverse and high-quality images \cite{saharia2022photorealistic, rombach2022high, ramesh2022hierarchical} guided by text prompts.  Existing diffusion-based editing methods  \cite{avrahami2022blended,avrahami2023blended,couairon2023diffedit,parmar2023zero,tumanyan2023plug,hertz2023prompttoprompt, ruiz2023dreambooth, meng2021sdedit} have predominantly addressed image-related tasks, and offer a wide range of functionalities such as object editing \cite{hertz2023prompttoprompt}, image inpainting \cite{avrahami2022blended}, personalized generation \cite{ruiz2023dreambooth}, and image-to-image translation \cite{meng2021sdedit}. These methods perform editing on images through a deterministic DDIM inversion process \cite{song2021denoising}. This involves conducting image-to-noise inversion and subsequently generating edited images from the noise based on the target text prompt. This new paradigm of interactive image editing and generation through textual prompts not only represents an exciting advancement but also lays the groundwork for a diverse spectrum of possibilities for content creation by nontechnical users.

Unfortunately, advances in image editing have not been translated into comparable success in video editing.
Prior research has been focused on text-to-video (T2V) generation \cite{ho2022imagen, singer2022make, ho2022video} by utilizing extensive text-video datasets such as WebVid-10M \cite{bain2021frozen}. However, these methods are prohibitively expensive and therefore unsuitable for broad use in content creation. 
Another line of research employs conventional video editing methods \cite{kasten2021layered,bar2022text2live,lee2023shape} such as keyframe selection \cite{jamrivska2019stylizing} or atlas editing \cite{kasten2021layered,bar2022text2live}. However, these methods are hampered by 
their dependence on time consuming processes like atlas learning  \cite{kasten2021layered, bar2022text2live, lee2023shape}, accurate keyframe selection  \cite{jamrivska2019stylizing}, and prompt-dependent tuning 
\cite{bar2022text2live, lee2023shape}. More recent works have  focused on leveraging pre-trained text-to-image (T2I) diffusion models  \cite{yang2023rerender, bar2022text2live, qi2023fatezero, wu2023tune, ceylan2023pix2video}. However, adapting pre-trained T2I models for video editing poses a significant challenge, as it is essential to ensure the temporal consistency within the edited video. While existing methods exhibit potential, they  face multiple limitations:  difficulty in editing long videos
~\cite{qi2023fatezero, ceylan2023pix2video}, challenges with complex edits such as shape \cite{yang2023rerender, geyer2023tokenflow, khachatryan2023text2video}, quality degradation in the presence of substantial motion \cite{khachatryan2023text2video}, demands significant processing time \cite{qi2023fatezero, wu2023tune}, or requires additional training \cite{liew2023magicedit, wu2023tune}.
\begin{figure}[t!]
    \centering
    \includegraphics[width=1.0\linewidth]{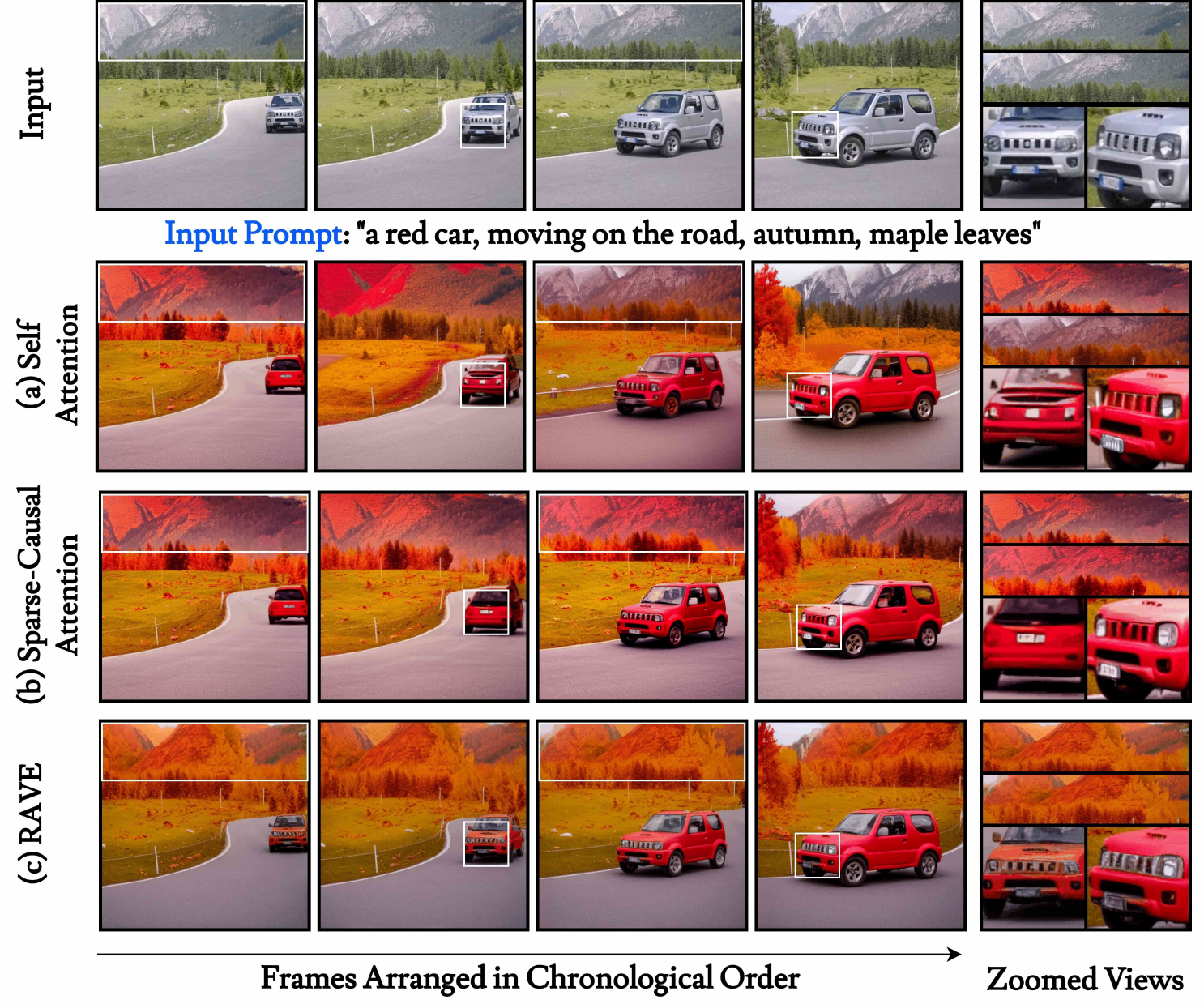}
    \caption{\textit{\textbf{Comparison with existing attention modules.}} The first row shows the frames of the input video, followed by subsequent rows presenting the edited frames under various attention settings, with our approach in the last row. The column on the right provides zoomed-in crops of mountains (from the 1\textsuperscript{st} and 3\textsuperscript{rd} frames) and the car's bumper (from the 2\textsuperscript{nd} and 4\textsuperscript{th} frames).} 
    \label{fig:attention}
\end{figure}
In this paper, we present \texttt{\methodName}, a novel text-guided zero-shot video editing approach that performs style, attribute, and shape editing on videos. It is compatible with any pre-trained T2I diffusion model, such as Stable Ddiffusion (SD) \cite{rombach2022high}, eliminating the need for training for every target prompt \cite{bar2022text2live, kasten2021layered, lee2023shape}, or user-specific masks \cite{avrahami2022blended, avrahami2023blended}, and also incorporates spatial guidance through ControlNet \cite{zhang2023adding}. Notably, RAVE is lightweight and efficient, achieving video edits at a rate $\sim$25\% faster than existing methods. 

Maintaining temporal consistency is crucial in video editing to ensure consistency across frames and to preserve the seamless movement of objects. Unlike previous studies \cite{qi2023fatezero, wu2023tune, ceylan2023pix2video} that depend on explicitly using multiple frames with attention methods like spatio-temporal attention or sparse causal attention for consistency~\cite{ceylan2023pix2video, wu2023tune}, RAVE employs a novel noise shuffling strategy that efficiently utilizes spatio-temporal interactions between frames. Our noise shuffling strategy inherently directs the model to perform spatio-temporal attention in a fraction of the usual time and memory requirements while still maintaining temporal consistency. Fig.~\ref{fig:attention} illustrates common attention strategies in video editing methods, highlighting their shortcomings and emphasizing the effectiveness of our noise shuffling approach. Using solely spatial self-attention can leverage pixel locations in feature maps to capture similar correlations, while cross-attention manages the correspondence between pixels and text prompts (see Fig.~\ref{fig:attention} (a)). However, as shown by the results, while the generated frames align with the text prompt in terms of motion and color style, they lack consistency due to the neglect of temporal context as seen in the inconsistencies in the background and the car's bumper. A straightforward approach would be to expand the self-attention module to include \textit{all} frames, but this presents a substantial computational challenge \cite{wu2023tune}. Processing all video frames simultaneously becomes impractical, especially for videos with a high frame count. Alternative spatio-temporal attention methods often use a sparse form of the mechanism, known as \textit{sparse-causal attention}~\cite{wu2023tune} where the attention matrices are calculated between the initial frame and preceding frames, or among selected keyframes \cite{wu2023tune, ceylan2023pix2video, hong2023improving, qi2023fatezero}. Although this method produces more consistent frames with reduced time complexity, its performance tends to decline in longer videos due to the diminishing temporal awareness. This decline is evident in Fig.~\ref{fig:attention} (b), as shown by the changes in the hood and the contrast in the mountains. In contrast, our strategy leverages the power of global spatio-temporal attention throughout \textit{all} frames, ensuring frame consistency even in long videos by performing a novel noise-shuffling operation at every diffusion step during sampling (see Fig.~\ref{fig:attention} (c)). This results in maintaining the reduced time complexity when editing long videos without sacrificing the quality or speed of image editing.
To summarize, our main contributions are as follows:

\begin{itemize}
    \item We present RAVE, a novel, training-free, zero-shot video editing framework that integrates with pre-trained T2I diffusion models, such as Stable Diffusion \cite{rombach2022high}. RAVE incorporates spatial guidance with ControlNet~\cite{zhang2023adding} to perform video style, attribute, and shape editing. 
   \item Our novel noise shuffling method harnesses spatio-temporal interactions between frames efficiently, resulting in temporally consistent videos $\sim$25\% faster than existing methods.

   \item We propose a novel video evaluation dataset that includes a diverse range of videos, from object-centric scenes to complex human motions like dancing and typing, as well as dynamic scenes featuring swimming fish and boats. Moreover, we share our source code publicly, hoping to promote further research.

\end{itemize}

%% file: sec_arxiv/2_related.tex
\section{Related Works}
\label{sec:related-works}

\paragraph{Text-driven image editing}
Methods such as Dream-Booth~\cite{ruiz2023dreambooth} and Textual Inversion~\cite{gal2023an} demonstrate diverse image generation through fine-tuning in a few-shot manner. UniTune~\cite{valevski2023unitune} and Imagic~\cite{kawar2023imagic}, both based on the Imagen~\cite{saharia2022photorealistic} model, exhibit strong editing performances. Recent training-free methods like Prompt-to-Prompt~\cite{hertz2023prompttoprompt}, DiffEdit~\cite{couairon2023diffedit}, Blended Diffusion~\cite{avrahami2022blended}, and Blended Latent Diffusion~\cite{avrahami2023blended} achieve local and detailed editing by leveraging attention properties.

\paragraph{Text-driven video editing with training} Text-guided video editing broadly falls into two categories. One recent work in the first category, Video Diffusion Models \cite{ho2022video}, uses a factorized space-time U-Net compatible with joint image-video training as its architecture. Dreamix \cite{molad2023dreamix} extends this with a mixed fine-tuning approach. In contrast, Make-A-Video \cite{singer2022make} reduces T2V training cost by eliminating the need for paired text-video data. MagicEdit \cite{liew2023magicedit} introduces a different learning setting, separating the learning of structure, motion, and content. However, these methods require extensive training on a large dataset, limiting scalability and generalizability.
Tune-A-Video \cite{wu2023tune} proposes tailored spatio-temporal attention as an extension to the T2I model, termed sparse causal attention. Shape-aware TLVE \cite{lee2023shape} suggests an atlas-learning-based approach enabling structure-editing. Edit-A-Video \cite{shin2023edit} introduces `sparse-causal blending' to reduce background inconsistency, utilizing Null-Text Inversion \cite{mokady2023null}, akin to Video-P2P \cite{liu2023video} and vid2vid-zero \cite{wang2023zero}. However, these methods are time-consuming, requiring optimization, which limits their real-world application capabilities.

\paragraph{Text-driven video editing without training}
Due to the optimization requirements of previous methods, recent studies emphasize zero-shot training-free approaches for practical applicability. Pix2Video \cite{ceylan2023pix2video} employs sparse-causal attention for temporal consistency along with latent guidance, using the predictions of the original images as a proxy at each denoising step. FateZero \cite{qi2023fatezero} uses attention features during inversion for spatio-temporal preservation and blending, claiming that those are better in preserving motion and structure compared to that of during sampling. Text2Video-Zero \cite{khachatryan2023text2video} synthesizes and edits videos with cross-frame attention, initial frame integration, and background smoothing. Rerender-A-Video \cite{yang2023rerender} employs hierarchical cross-frame constraints for temporal consistency, while TokenFlow \cite{geyer2023tokenflow} focuses on feature-level smoothing to reduce the effects of flickering.

Nevertheless, FateZero~\cite{qi2023fatezero}, similar to Pix2Video~\cite{ceylan2023pix2video}, requires source prompts as it is built on the Prompt-to-Prompt~\cite{hertz2023prompttoprompt} editing technique. This method necessitates specific types of prompts on the source prompt, thereby limiting editing diversity. Additionally, both are constrained to shorter clips due to memory limitations. Text2Video-Zero \cite{khachatryan2023text2video} and Rerender \cite{yang2023rerender} heavily rely on off-the-shelf methods and optical flow, limiting consistency over longer videos. TokenFlow \cite{geyer2023tokenflow} faces structure editing limitations, dependent on inversion quality, and feature-level smoothing causes blurring in the edited image. As opposed to our method, previous approaches require time-intensive operations that slow down inference speed, restricting applicability to longer videos.
\begin{figure*}
    \centering
\includegraphics[width=1.0\textwidth]{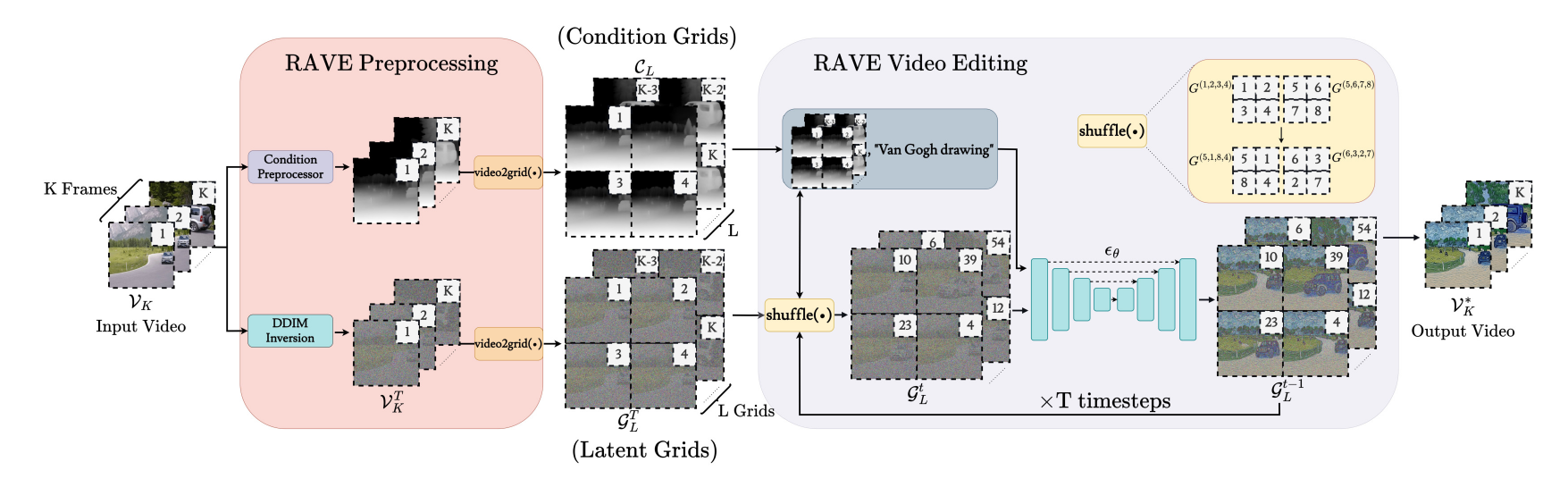}
    \caption{\textit{\textbf{An illustration of RAVE.}} Our process begins by performing a DDIM inversion with the pre-trained T2I model and condition extraction with an off-the-shelf condition preprocessor applied to the input video ($\mathcal{V}_K$).
    These conditions are subsequently input into ControlNet. 
In the RAVE video editing process, diffusion denoising is performed for T timesteps using condition grids ($\mathcal{C}_L$), latent grids ($\mathcal{G}_L^t$), and the target text prompt as input for ControlNet. Random shuffling is applied to the latent grids ($\mathcal{G}_L^t$) and condition grids ($\mathcal{C}_L$) at each denoising step. After T timesteps, the latent grids are rearranged, and the final output video ($\mathcal{V}_K^*$) is obtained.}
    \label{fig:framework}
    \vspace{-1.5em}
\end{figure*}

%% file: sec_arxiv/3_methodology.tex
\section{Methodology}
\label{sec:methodology}
Consider a video, denoted as $\mathcal{V}_K=\{\mathcal{I}_1, \ldots, \mathcal{I}_K\}$, consisting of $K$ individual frames. Each frame, represented as $\mathcal{I}_k \in \mathbb{R}^{W\times H\times C}$ corresponds to the $k\textsuperscript{th}$ position in the video sequence where $W$, $H$, and $C$ specify the width, height, and color channels, respectively. Suppose we have a text prompt $\mathcal{P}$, detailing a specific edit to be made to the video. Our goal is to generate a modified video, $\mathcal{V}^{*}_K$, where the original video is transformed based on the changes outlined in $\mathcal{P}$. For example, imagine a video of a monkey sitting on the grass and eating food (refer to Fig.~\ref{fig:main_results}). If given a prompt such as \textit{``a teddy bear is eating an apple"}, our task would be to transform the monkey and food into a teddy bear and apple, while preserving its original motion and semantic layout. As we achieve this transformation using only publicly available T2I models combined with a single video-text pair, we consider our approach as \textit{zero-shot text-guided video editing}.

\subsection{Preliminaries}
\label{subsec:preliminaries}

\paragraph{Latent diffusion models (LDMs)} 
Denoising diffusion probabilistic models (DDPM) are computationally intensive due to repetitive denoising steps in pixel space. Addressing this, LDMs~\cite{rombach2022high} introduce two key extensions: (i) The diffusion process operates in a lower-dimensional space by initially encoding the image using a pre-trained autoencoder, and (ii) improved controllability is achieved through conditions during training. LDMs typically employ a U-Net \cite{ronneberger2015u} as the denoising model, featuring skip connections between decoder and encoder levels. The U-Net consists of stacked 2D convolutional residual blocks and transformer blocks, each with a spatial \textit{self-attention} (S-A) layer capturing spatial correlations, a \textit{cross-attention} (C-A) layer interacting with conditional inputs (\eg tokens in the text prompt), and a \textit{feed-forward layer} processing the output for compatibility with the subsequent block.

\paragraph{Denoising diffusion implicit models (DDIM)}
To expedite the sampling process of diffusion models, DDIM~\cite{song2021denoising} generalizes the Markov structure of DDPM~\cite{ho2020denoising} to a non-Markov setting and allows us to reach the clear sample with less steps. 
That allows us to obtain the corresponding latent from the original latent,
$z_0$, for each timestep $t$, $\{{z}_t\}_{t=1}^{t=T}$.
For image editing, one needs to map the given image to its inverse counterpart.
Therefore, this technique is widely utilized in various image/video editing approaches with diffusion models.

\paragraph{ControlNet}
 To enhance flexibility and controllability of the video editing process, our method incorporates ControlNet \cite{zhang2023adding} for structural guidance. ControlNet introduces a control mechanism for diffusion models by integrating a weight locking strategy, allowing for additional conditions beyond the text prompt $\mathcal{P}$, such as line art, depth, pose, and surface normals. 
 This feature facilitates more controlled generation, particularly in the structure of images, offering an advantage over traditional LDMs.

\subsection{Our approach}
\label{subsec:consistency}

\paragraph{Grid trick} The grid trick, also known as \textit{character sheet}, is a popular hack in SD \cite{rombach2022high} art community. It is frequently used by enthusiasts and hackers for avatar design and image stylization \cite{gridtrick}.  This approach allows for applying a consistent style to all images in a grid using a T2I model, as shown in Fig.~\ref{fig:grid_example}. The effectiveness of this simple method in producing multiple images with consistent style lies in how the diffusion model performs editing of a single grid. It treats all frames collectively as a single image through its convolution-based residual blocks and transformer blocks. The self-attention layers in the transformer blocks are especially crucial for enhancing temporal consistency across frames in a grid. Simultaneously, the convolution layers within the residual block are responsible for capturing both temporal connections and spatial correlations. Note that, to the best of our knowledge, we are the first to systematically utilize this trick in image/video editing research.

\begin{figure}[t!]
    \centering
\includegraphics[width=1.0\linewidth]{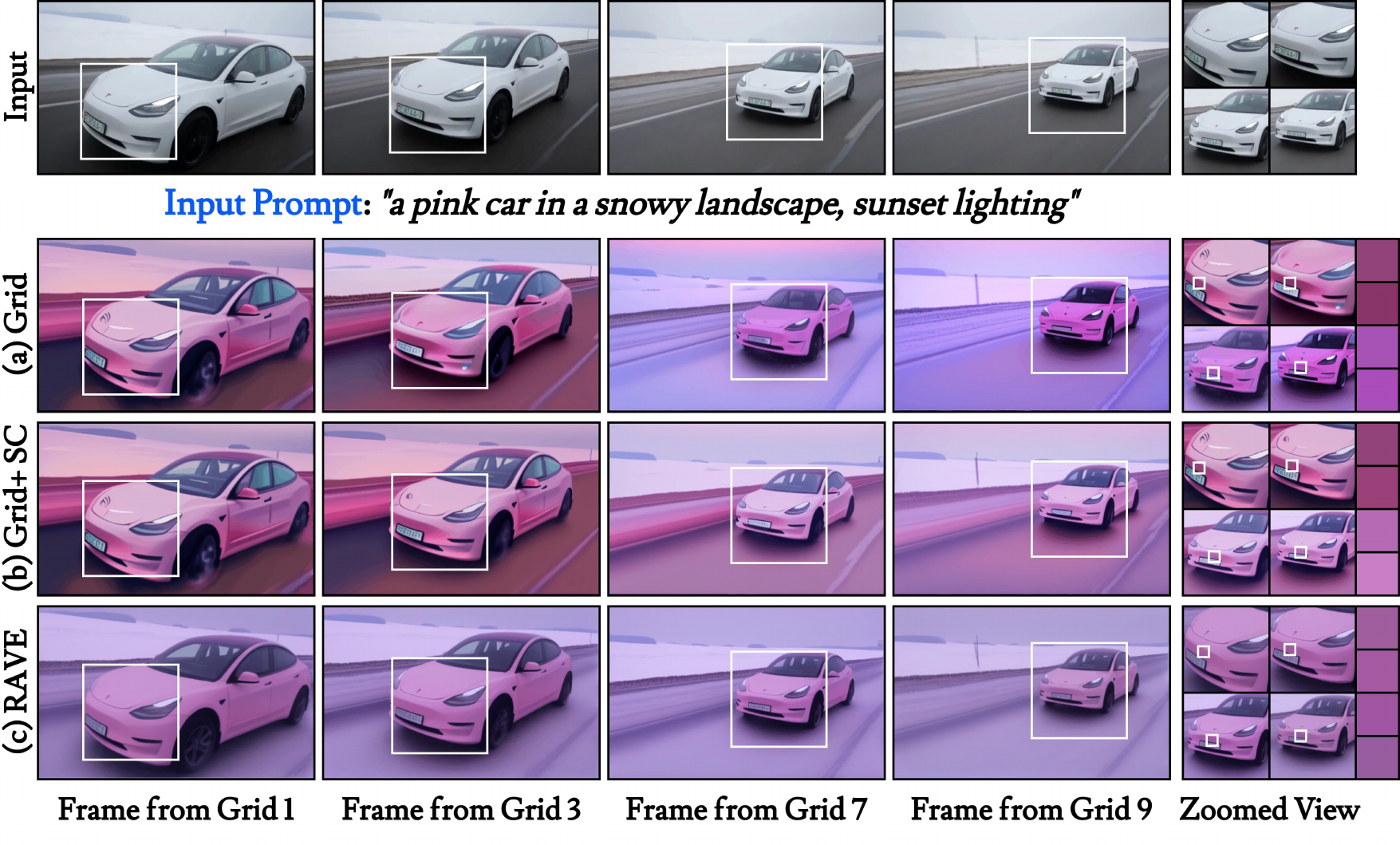}
\caption{\textit{\textbf{Consistency across grids.}}  Editing results are shown for (a) processing grids independently, (b) adapting sparse-causal   attention using grids, and (c) applying RAVE. The rightmost column features a close-up of the car's front, highlighting temporal color changes per approach. RAVE produces consistent patches in all grids while other methods struggle with consistency.}
\label{fig:consistency_grids}
\end{figure}

\paragraph{Grid trick for video editing} A straightforward adaptation of this trick for video editing involves transforming the input video into a grid layout and then feeding this grid to T2I editing tools such as ControlNet. Subsequently, the modified grid can be unfolded into separate frames and sequenced consecutively to produce the final video. More formally, assume that each frame in the video is arranged into a $N=n\times m$ grid with $n$ rows and $m$ columns, denoted as $G^{(1,\cdots, N)}\in\mathbb{R}^{(W \cdot m) \times (H \cdot n) \times 3}$, where $H$ and $W$ represent the height and width of the frames in the video, respectively, and superscript $(1,\cdots,N)$ denotes the indices of frames included in the grid. During the editing process of a single grid $G^{(1,\cdots, N)}$, the diffusion model processes $N$ frames holistically as a single image by fostering a \textit{smooth transition} between the latent vectors,  $\{{z}_t^0, \ldots, {z}_t^n\}_{t=1}^{t=T}$, $z_t^i$ denoting the corresponding latent region of $i^{th}$ frame within the grid at timestep $t$. This method achieves consistent frames in short sequences as can be seen from Fig \ref{fig:consistency_grids} (a), where the frames from grid 1 and grid 3 share the same style. However, it encounters challenges in maintaining this consistency in longer videos, even with the same text prompt. This limitation arises due to GPU memory constraints, restricting the number of frames that can fit within a grid. For example, if a GPU can only process a maximum of a $3\times3$ grid, equivalent to 9 frames, videos exceeding 9 frames need to be divided into multiple grids, each processed separately.
Consequently, the sampling trajectories for each grid diverge, leading to internal consistency within each grid but noticeable style differences between grids, as evident in Fig.~\ref{fig:consistency_grids} (a), where the style of frames from separate grids differs.   

\paragraph{Noise shuffling}
While the grid technique enables consistent editing, ensuring consistency \textit{across} multiple grids remains a challenge. One could modify well-known attention mechanisms, like sparse-causal attention, for the grid structure. In this adaptation, attention is shifted from focusing on the initial frame and the previous frame to the \textit{initial grid} and the \textit{previous grid}.  However, this approach can still face difficulties in maintaining consistency with longer videos. For instance, Fig.~\ref{fig:consistency_grids} (b) demonstrates that although sparse-causal attention works effectively for frames from grid 1 and grid 3, it encounters challenges in maintaining consistency as the video progresses. In order to  achieve consistency across grids, we introduce a novel \textit{noise shuffling} approach that ensures consistent styling across independent grids (see Fig.~\ref{fig:consistency_grids} (c)). 
\

Consider the \textit{video2grid}$(\cdot, \cdot)$ function that accepts an input video and a grid size $N = n \times m$, and generates a series of grids while preserving the original frame order. For instance, when the original video $\mathcal{V}_K$ is used as input, we can produce a set of $L$ grids as:
\begin{align}
   \textit{video2grid}(\mathcal{V}_K, N) &=  \{G_1^{(1,\cdots, N)}, \ldots, G_L^{ (P, \cdots, K)} \}\\
   &= \mathcal{G}_L
\end{align} 
where $P=K-(N+1)$.
 In this context, $L$ represents the total number of grids, calculated as $L=K/N$, assuming that $K$ is divisible by $N$. 

 Rather than processing each grid in $\mathcal{G}_L$ sequentially to produce the final video output, which gives undesirable results (as depicted in Fig.~\ref{fig:consistency_grids} (a)), our approach entails randomly shuffling the order of frames at each timestep of the sampling process and allocating frames to different grids in a random order. More formally, starting with the input grids generated by the \textit{video2grid}$(\cdot, \cdot)$ operation, we proceed as follows:
\begin{align}
       \textit{shuffle}(\mathcal{G}_L) = \{G_1^{(i,\cdots, j)}, \ldots, G_L^{(k, \cdots, l)} \}
\end{align}
where $(i,\cdots, j),\cdots,  (k, \cdots, l) \in \{1,\cdots,K\}$ are randomly chosen non-repeating indices. In other words, this function randomly rearranges the locations of the frames in a given set of grids. Through the random formation of grids from latents arranged in a random order, we encourage spatio-temporal interaction between each frame throughout the denoising process. This interaction involves convolutional layers, which play a role in globally smoothing the latent vectors, thereby reducing pixel shifts that can lead to flickering. Furthermore, self-attention layers ensure that features of each patch from every frame are influenced by patches of other frames at each step, guaranteeing the application of the same style globally. This enhancement contributes to temporal consistency, regardless of the video size since our approach does not require extra memory requirement proportional to the length of the video. The complete structure of our approach is depicted in Fig.~\ref{fig:framework}.

%% file: sec_arxiv/4_dataset.tex
\section{Dataset}

The availability of datasets featuring a wide range of editing tasks and motions for the evaluation of zero-shot text-guided video editing tasks is limited (see \ref{sec:dataset_details} for a detailed discussion). To foster standardized evaluations in future video editing research, we have curated a dataset comprising 186 text-video pairs from diverse sources, including Pexels \cite{pexels}, Pixabay \cite{pixabay}, DAVIS \cite{perazzi2016benchmark}, and Internet videos used by previous approaches. The prompts are obtained from ChatGPT, drawn from previous approaches, or contributed by the authors. 

\paragraph{Length and Resolution} 
The dataset categorizes video lengths into three segments—8, 36, and 90 frames—with 10, 15, and 6 videos respectively, to analyze the impact of duration on video editing methods. Note that the existing methods can handle only up to a certain number of frames;  for instance FLATTEN~\cite{cong2023flatten} can handle up to 27 frames on RTX4090\footnote{\href{https://openreview.net/forum?id=JgqftqZQZ7}{FLATTEN} Last accessed: 2023-11-26.} and FateZero~\cite{qi2023fatezero} is up to 45 frames on A40 with their official repositories. Hence, assessing video editing methods for longer video lengths is crucial. We also use resolutions of $512\times 320$ or $512\times 256$ for rectangular, and $512\times512$ for square videos.

\begin{figure}[t!]
    \centering
\includegraphics[width=1.0\linewidth]{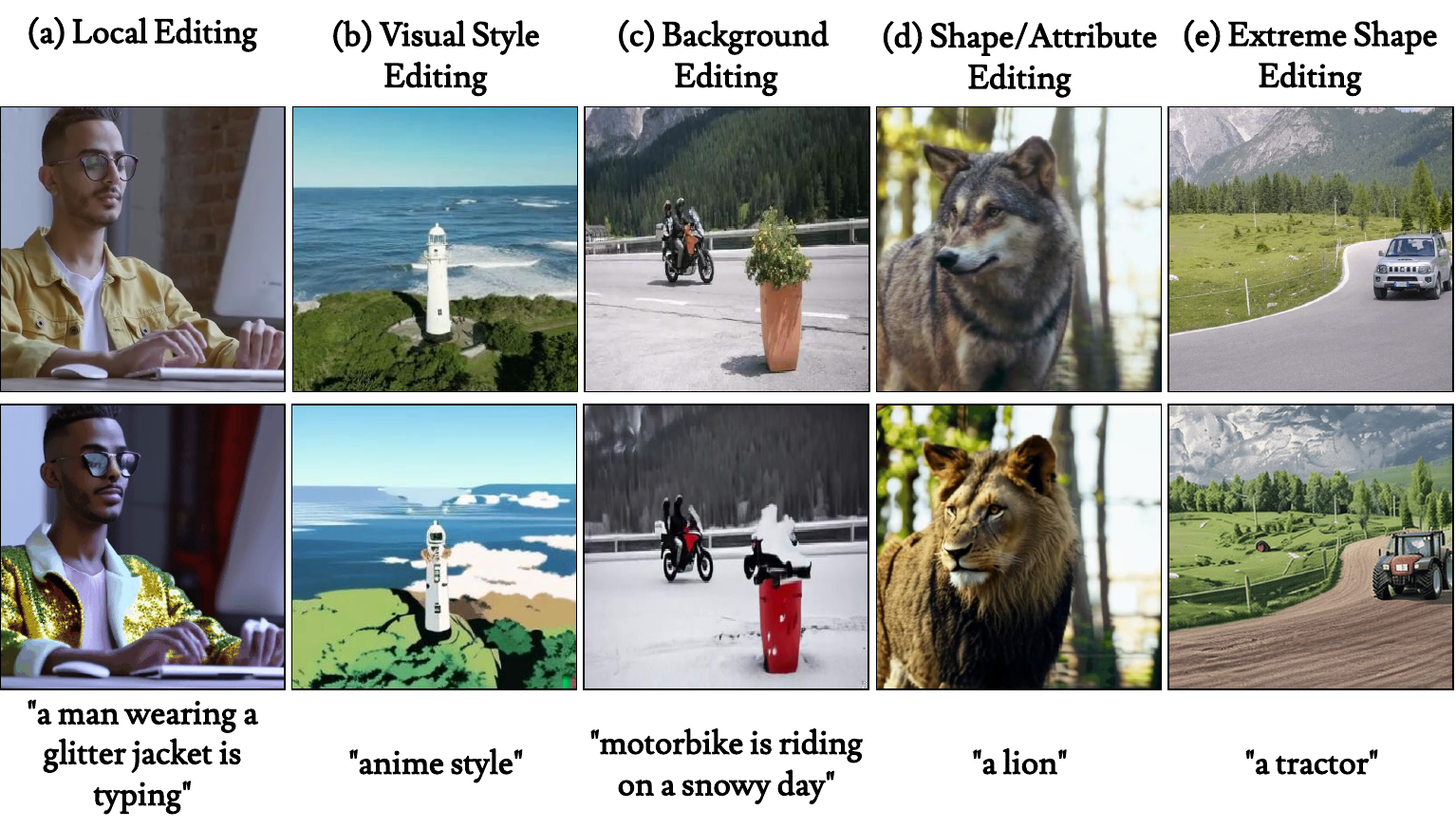}
    \caption{\textbf{\textit{Types of edits in our dataset.}} }
\label{fig:style-supp}
\end{figure}

\paragraph{Types of Edits}
We broadly classify editing types into \textbf{style editing} and \textbf{shape editing} (see Fig. \ref{fig:style-supp}). Style editing is divided into: 
\begin{itemize}
    \item \textbf{Local editing} for localized changes (\eg jacket color),
    \item \textbf{Visual style editing} for artistic style changes (\eg `watercolor style'),
    \item \textbf{Background editing} for background or setting changes (\eg beach background).
\end{itemize}
Furthermore, shape editing is divided into two types:
\begin{itemize}
    \item \textbf{Shape/attribute editing} focusing on altering an object's shape or attributes (e.g., wolf to cat), 
    \item \textbf{Extreme shape editing} for major transformations of an object's shape or nature (e.g.,  car to tractor).
\end{itemize}

 \begin{figure*}
    \centering
\includegraphics[width=1.0\linewidth]{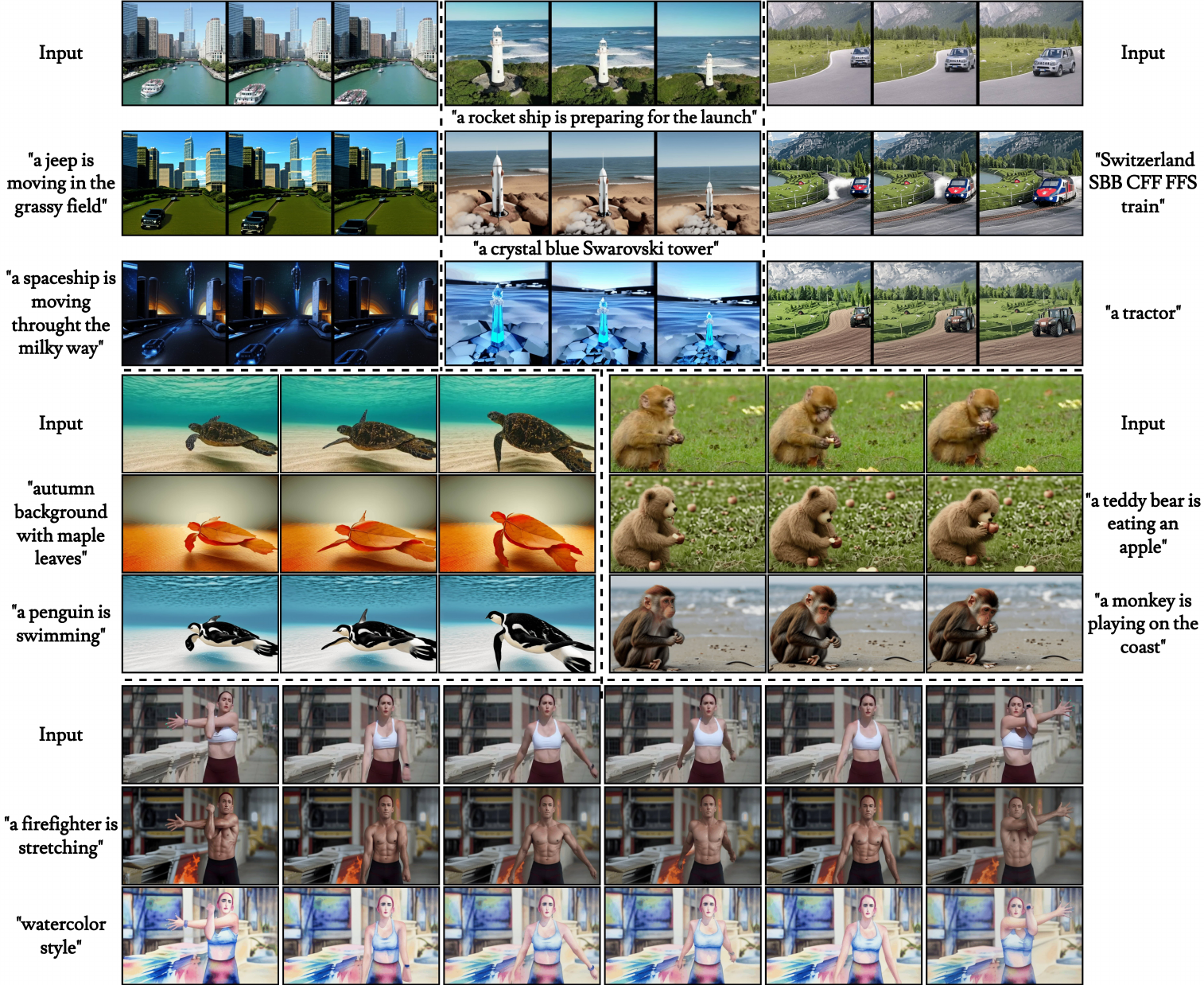}
\caption{\textit{\textbf{Qualitative results with RAVE.}} Our method is capable of performing a variety of video editing tasks, including style editing, local attribute editing, and shape editing (top right). It can edit videos featuring moving objects, such as turtles or boats, as well as videos with complex human movements, like a woman stretching (bottom row).
    Best viewed in zoomed-in. For more, please see our \href{https://rave-video.github.io}{webpage}.}
\label{fig:main_results}
\vspace{-1.5em}
\end{figure*}

\paragraph{Motion} We assess the robustness of baseline approaches across different types of motions.

\begin{itemize}
    \item \textbf{Exo-motion}, where only the object within the video is in motion, while background stays the same,
    \item \textbf{Ego-motion}, involving scenarios where the camera itself is in motion.

    \item \textbf{Ego-exo motion}, combines both camera and object movement in the videos, 
    \item \textbf{Occlusion}, incorporates scenarios where objects are partially or fully hidden from view, 
    \item \textbf{Multiple objects with appearances/disappearances}, features videos with multiple objects, some appearing and disappearing throughout the video. 
\end{itemize}

By considering these diverse types of motions, we can thoroughly evaluate the robustness of the approaches in addressing the complexities of real-world scenarios. This assessment provides valuable insights into the methods' adaptability and effectiveness across a wide range of dynamic situations in video editing.

\paragraph{Diversity of Text Prompts} Our dataset involves employing varied levels of detail in our text prompts. This includes  examples such as `\textit{a zombie}' and more elaborate descriptions like `\textit{Soft, blended colors and visible brushstrokes make the scene appear as if painted with watercolors}'.

%% file: sec_arxiv/5_exp.tex
\section{Experimentation}
\label{sec:exp}

\paragraph{Baselines}

We conduct a thorough quantitative and qualitative comparison with concurrent baselines, using their official implementations. Rerender \cite{yang2023rerender} relies on optical flow, Text2Video-Zero \cite{khachatryan2023text2video} employs random warping and optional background smoothing, limiting its effectiveness in videos with background motions. TokenFlow \cite{geyer2023tokenflow} applies feature-level smoothing but has constraints in shape-aware editing. FateZero \cite{qi2023fatezero} is excluded from quantitative comparison due to its requirement for target prompts to be in a specific format with the source prompts, necessitating manually crafted prompts. Additionally, both FateZero~\cite{qi2023fatezero} and Pix2Video~\cite{ceylan2023pix2video} face GPU usage limits, which cannot process videos for more than 22 and 45 frames for Pix2Video and FateZero with their official repositories, respectively. We use official repositories and default parameters of the baselines\footnote{We use the suggested parameters provided by the authors after reaching out to them and default parameters otherwise.}, and report the results of RAVE and RAVE without shuffling as an ablation.

\paragraph{Implementation details}
We use Stable Diffusion (SD) 1.5 from the official Huggingface repo for all baselines in any type of comparison. For qualitative assessment of our approach, the Realistic Vision V5.1 model from the CivitAI~\cite{civitai} repo is employed.  All approaches use 50 steps of DDIM inversion and sampling along with a fixed classifier-free guidance scale of 7.5. RAVE employs a $2\times 2$ grid size for videos with a length of 8 frames and a $3\times 3$ grid size for the rest. In all comparisons, including quantitative, qualitative and user study comparison, we applied depth-conditioned ControlNet for our method as well as for Pix2Video, Text2Video-Zero, and Rerender. TokenFlow utilizes Plug-and-Play \cite{tumanyan2023plug} for image editing method, therefore we use Plug-and-Play in their method during comparison. We use the official repositories of the previous approaches. We do not use any negative or positive prompts in RAVE. Also for a fair comparison, no positive prompts or negative prompts are used for other methods as well. We run all experiments on a single A40 GPU.

\begin{figure*}    \centering\includegraphics[width=1.0\linewidth]{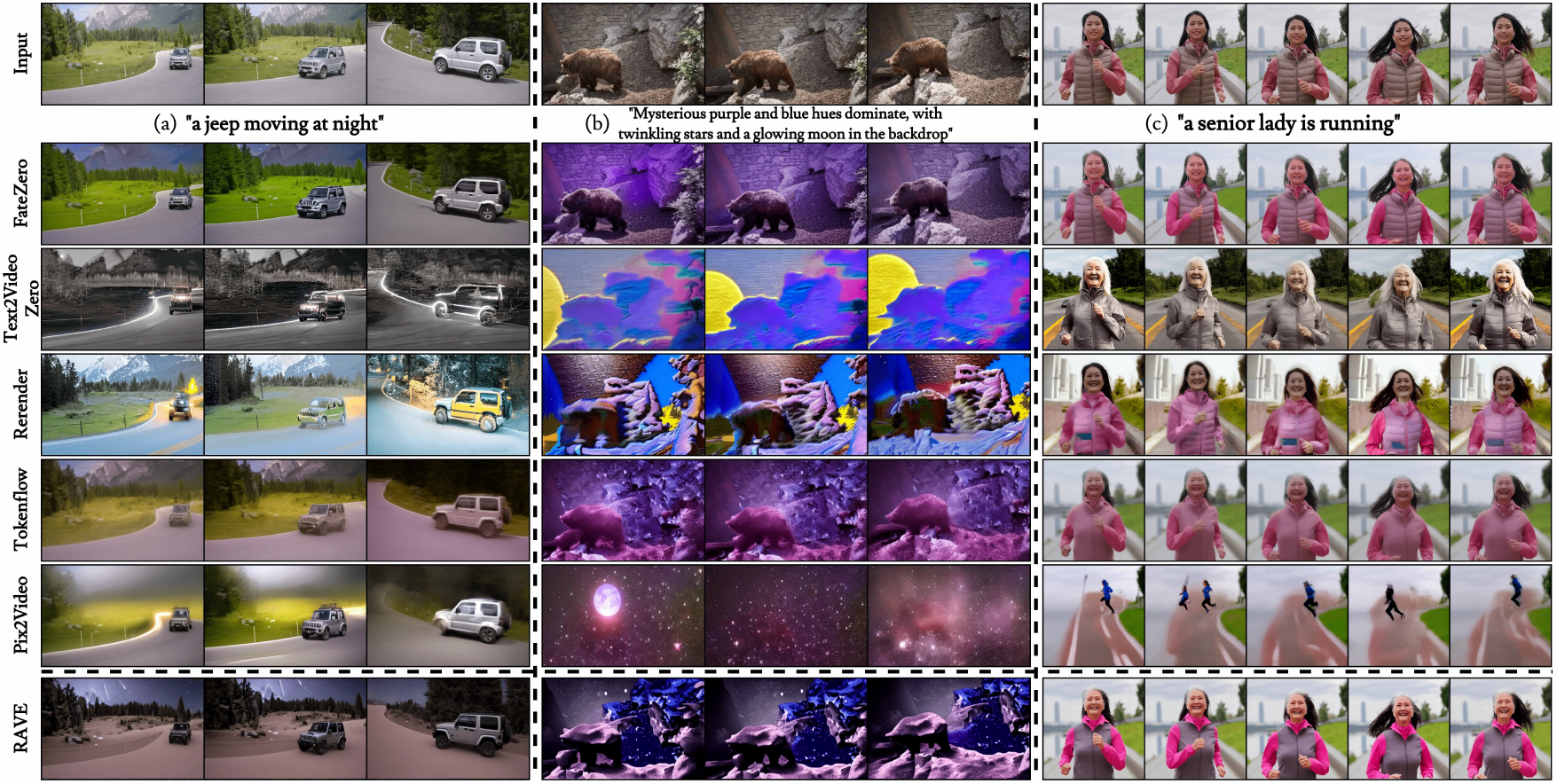}   \caption{\textit{\textbf{Qualitative comparison.}} We present a comparison of approaches on videos featuring diverse motions and objects on short and long videos. Videos in columns (a) and (b) have a total of 36 frames, while video in (c) has 90 frames. Due to their excessive GPU requirements, Pix2Video and FateZero are executed on shorter clips. Results are best viewed in zoomed-in. For more, please see \href{https://rave-video.github.io}{https://rave-video.github.io}.}
\label{fig:main_comparison}
\vspace{-1.5em}
\end{figure*}

\subsection{Evaluation}
\label{sec:quantitative}

\paragraph{Qualitative assessment} 
Fig.~\ref{fig:teaser} and Fig.~\ref{fig:main_results} illustrate some examples of video editing using RAVE. We demonstrate style editing (\eg watercolor style), complex shape editing (\eg transforming a car into a tractor), and both applied simultaneously (\eg `a jeep is moving in the grassy field'). We use videos featuring backgrounds in motion (\eg top right video) and different objects (\eg turtle, human, watchtower) engaged in various activities (\eg swimming, eating, stretching).
Moreover, we present a qualitative comparison in Fig.~\ref{fig:main_comparison} with baseline methods. Notice that FateZero struggles to edit non-local attributes, as evident with twinkling stars in Fig.~\ref{fig:main_comparison} (b) since the quality of edit depends on Prompt-to-Prompt~\cite{hertz2023prompttoprompt} which heavily relies on the chosen parameters and source prompt. Text2Video-Zero appears effective with videos featuring constant backgrounds, despite a color change in the woman's jacket Fig.~\ref{fig:main_comparison} (c); however, it encounters challenges when there is motion in the background, such as the trees in Fig.~\ref{fig:main_comparison} (a). Rerender heavily utilizes optical flow along with keyframe propagation, however these are not enough to perform editing while keeping style over time (observe the color change of the car in Fig.~\ref{fig:main_comparison} (a)). Although TokenFlow excels at maintaining structural consistency over time, supported empirically by quantitative results (WarpSSIM), it experiences overall blurring in the entire image due to `feature-level smoothing' being applied. Pix2Video's edits deviate significantly from the source video (note the woman in Fig.~\ref{fig:main_comparison} (c) and the bear in Fig.~\ref{fig:main_comparison} (b)), indicating that latent guidance alone and depth guidance solely are insufficient for consistency in long videos. In contrast, our method preserves global consistency, aligns with the text prompt, and maintains  editing quality—all accomplished in less time compared to previous approaches (see Table~\ref{tab:quant} last column).

\paragraph{Quantitative evaluation} Following prior work~\cite{geyer2023tokenflow, ceylan2023pix2video, yang2023rerender}, we assess the temporal consistency of editing at a coarse level by measuring the average pairwise similarities (\textbf{CLIP-F}) of each frame using CLIP~\cite{radford2021learning} and structural consistency (\textbf{WarpSSIM}) by calculating the average SSIM~\cite{wang2004image} score between the warped edited video (RAFT~\cite{teed2020raft} is used to obtain the optical flow of the source video) and the edited video. Additionally, we evaluate the textual alignment of the editing, given a prompt, by calculating the average distance between the CLIP embedding of the text prompt and all frames (\textbf{CLIP-T}). Note that these metrics alone do not provide a comprehensive evaluation since CLIP-T is unrelated to consistency, and WarpSSIM and CLIP-F can yield high scores even when the video is not edited at all. Therefore, following  \cite{cong2023flatten}, we adopt \textbf{Q\textsubscript{edit}} = WarpSSIM $\cdot$ CLIP-T for a more holistic assessment. The results are shown in Table~\ref{tab:quant}.  Quantitative evaluation is performed on our dataset, consisting of 10, 15, and 6 videos with lengths of 8, 36, and 90 frames, respectively. Each video is edited with 4 style and 2 shape prompts, resulting in 186 text-video pairs. Our performance surpasses previous approaches in both frame consistency and textual alignment. While TokenFlow demonstrates slightly greater effectiveness for shorter frames in terms of structural consistency (WarpSSIM), our superiority becomes evident for longer videos, specifically at a frame length of 90. Moreover, when assessing comprehensive editing capability with Q\textsubscript{edit}, we outperform other methods in all video lengths. Additionally, there is a significant increase in the CLIP-F score, particularly in longer videos (36 and 90 frames), when comparing RAVE w/o shuffling and RAVE. This emphasizes the importance of shuffling in preserving consistency.

\begin{table*}[]
\centering
\caption{\textit{\textbf{Quantitative comparison.}} 
CLIP-F, WarpSSIM, CLIP-T, and Q\textsubscript{edit} metrics are reported individually on videos of 8, 36, and 90 frames. The user study section reports the frequency of each method chosen among the top two edits for General Editing (Q1 (GE)), Temporal Consistency (Q2 (TC)), and Textual Alignment (Q3 (TA)). The last column presents video-editing runtime in `minutes:seconds' format for 90 frames for the entire pipeline, including preprocessing and editing stages (parentheses indicate runtime w/o preprocessing). '-' denotes methods that cannot be measured due to excessive memory requirements, while 'N/A' indicates that the value is not available.}
\resizebox{1.0\textwidth}{!}{
\begin{tabular}{|l|ccc|ccc|ccc|ccc|ccc|c|}
\toprule
\multicolumn{1}{|c|}{\textbf{Method}}& \multicolumn{3}{c|}{\textbf{CLIP-F} ($\times 10^{-2}$) $\uparrow$} & \multicolumn{3}{c|}{\textbf{WarpSSIM} ($\times 10^{-2}$) $\uparrow$} & \multicolumn{3}{c|}{\textbf{CLIP-T} ($\times 10^{-2}$) $\uparrow$} & \multicolumn{3}{c|}{\textbf{Q\textsubscript{edit}} ($\times 10^{-5}$)  $\uparrow$}  & \multicolumn{3}{c|}{\textbf{User Study}  $\uparrow$} & \multicolumn{1}{c|}{\textbf{Runtime} $\downarrow$} 
\\ 
 & 8-frames & 36-frames & 90-frames & 8-frames & 36-frames & 90-frames & 8-frames & 36-frames & 90-frames & 8-frames & 36-frames & 90-frames  &  Q1 (GE) & Q2 (TC) & Q3 (TA) & 90-frames \\
\midrule
\textbf{Text2Video-Zero} & \multicolumn{1}{c|}{95.49} & \multicolumn{1}{c|}{92.89} & 94.35 & \multicolumn{1}{c|}{67.97} & \multicolumn{1}{c|}{36.65} & 71.57 & \multicolumn{1}{c|}{\textit{29.46}} & \multicolumn{1}{c|}{29.42} & \textit{29.73} & \multicolumn{1}{c|}{\cellcolor{green!25}20.02} & \multicolumn{1}{c|}{\cellcolor{green!25}10.78} & \cellcolor{green!25}21.27 & \textit{47.95\%} & 24.87\% & \textit{52.56\%} & 5:33 \\   
   
\textbf{Rerender} & \multicolumn{1}{c|}{92.87} & \multicolumn{1}{c|}{89.71} & 90.63 & \multicolumn{1}{c|}{68.57} & \multicolumn{1}{c|}{44.54} & 74.56 & \multicolumn{1}{c|}{25.65} & \multicolumn{1}{c|}{27.42} & 27.55 & \multicolumn{1}{c|}{\cellcolor{green!25}17.66} & \multicolumn{1}{c|}{\cellcolor{green!25}12.24} & \cellcolor{green!25}20.51 & {17.44\%} & {23.33\%} &  {17.18\%} & 5:24 \\     
\textbf{TokenFlow} & \multicolumn{1}{c|}{\textit{95.80}} & \multicolumn{1}{c|}{\textit{93.17}} & \textit{95.92} & \multicolumn{1}{c|}{\textbf{74.03}} & \multicolumn{1}{c|}{\textbf{50.97}} & \textit{80.40} & \multicolumn{1}{c|}{28.27} & \multicolumn{1}{c|}{28.29} & 29.53 & \multicolumn{1}{c|}{\cellcolor{green!25}\textit{20.92}} & \multicolumn{1}{c|}{\cellcolor{green!25}\textit{14.41}} & \cellcolor{green!25}\textit{23.74} & {44.10\%} & \textit{68.97\%} & {43.59\%}  & 5:24 (4.14) \\   
\textbf{Pix2Video} & \multicolumn{1}{c|}{89.96} & \multicolumn{1}{c|}{-} & - & \multicolumn{1}{c|}{24.78} & \multicolumn{1}{c|}{-} & - & \multicolumn{1}{c|}{28.01} & \multicolumn{1}{c|}{-} & - & \multicolumn{1}{c|}{\cellcolor{green!25}5.61} & \multicolumn{1}{c|}{\cellcolor{green!25}-} & \cellcolor{green!25}- & N/A & N/A & N/A &  -  \\  
\midrule
\textbf{RAVE - w/o shuffle} & \multicolumn{1}{c|}{93.98} & \multicolumn{1}{c|}{89.90} & 92.49  & \multicolumn{1}{c|}{\textit{71.78}} & \multicolumn{1}{c|}{47.26} & 76.58 & \multicolumn{1}{c|}{28.78} & \multicolumn{1}{c|}{\textit{29.49}} & 29.71 & \multicolumn{1}{c|}{\cellcolor{green!25}20.66} & \multicolumn{1}{c|}{\cellcolor{green!25}13.94} & \cellcolor{green!25}22.76 & N/A  & N/A &  N/A & N/A \\     
\textbf{RAVE} & \multicolumn{1}{c|}{\textbf{95.95}} & \multicolumn{1}{c|}{\textbf{93.18}} & \textbf{95.99} & \multicolumn{1}{c|}{71.44} & \multicolumn{1}{c|}{\textit{48.81}} & \textbf{80.51} & \multicolumn{1}{c|}{\textbf{29.51}} & \multicolumn{1}{c|}{\textbf{29.93}} & \textbf{29.76} & \multicolumn{1}{c|}{\cellcolor{green!25}\textbf{21.08}} & \multicolumn{1}{c|}{\cellcolor{green!25}\textbf{14.60}} & \cellcolor{green!25}\textbf{23.95}& \textbf{90.51\%} & \textbf{82.82\%} & \textbf{86.67\%} &  \textbf{4:28} (\textbf{3:13}) \\    
\bottomrule
\end{tabular}}
\label{tab:quant}
\vspace{-1.5em}
\end{table*}
\paragraph{User study} While quantitative metrics offer a fair comparison among prior methods, assessing edit quality and temporal consistency remains challenging. Hence, we conduct a user study, posing questions (Q1: General Editing (GE), Q2: Temporal Consistency (TC), Q3: Textual Alignment (TA)) to 130 anonymous participants on a crowdsourcing platform, Prolific~\cite{prolific}, for randomly selected 23 video-text pairs. Our survey results demonstrate superiority over previous methods in both aspects. TokenFlow, as anticipated due to its success in WarpSSIM score, secures the second-best results for temporal consistency.

\paragraph{Runtime} 
We conduct a runtime comparison (shown in rightmost column in Table~\ref{tab:quant}). Our results indicate that RAVE is  $\sim$1 minute faster than the closest competitor, TokenFlow, on average for editing a single video of 90 frames, making it the preferred choice for fast video editing. Note that both our method and TokenFlow require preprocessing only once for a given video, regardless of the text prompt. In contrast, Text2Video-Zero uses random noises, and Rerender performs inversion only on the keyframes. Runtimes without preprocessing durations are shared in Table~\ref{tab:quant}, rightmost column in parenthesis.

\begin{figure}[h!]
    \centering
\includegraphics[width=0.99\linewidth]{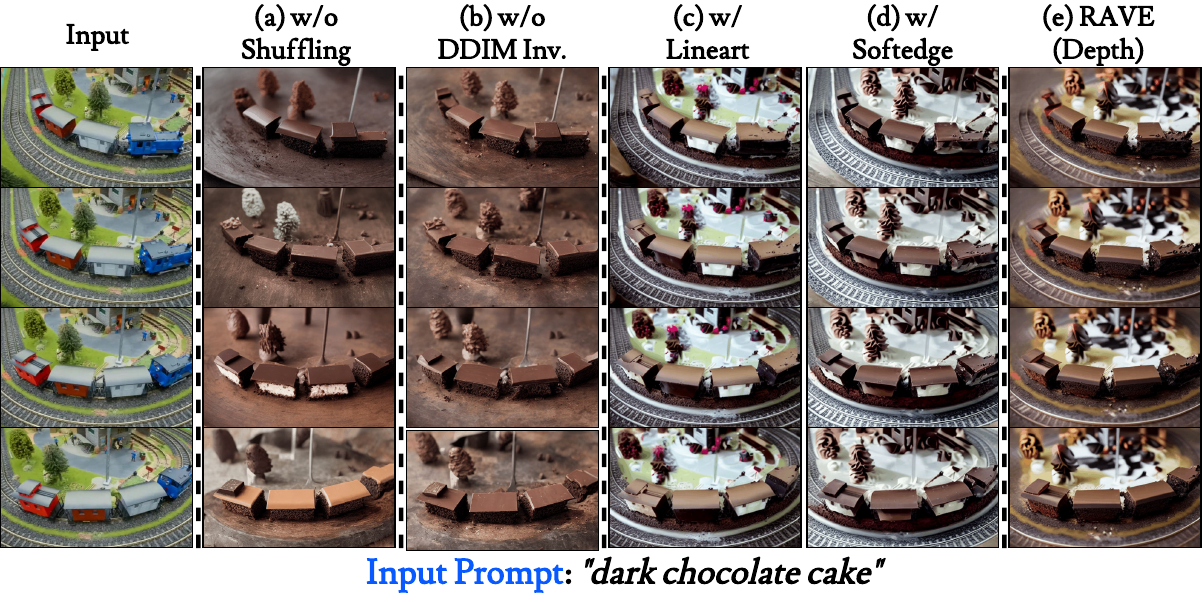}
    \caption{\textbf{\textit{Ablation study.}} We ablate the following components: (a) shuffling, (b) DDIM inversion, and employ different conditions: (c) lineart and (d) softedge (e) depth.}
\label{fig:ablation_main}
\end{figure}

\paragraph{Ablation study} We conduct an ablation study by separately ablating `shuffling', `DDIM inversion', and ControlNet conditions (lineart, softedge and depth (RAVE)) in our framework, as illustrated in Fig.~\ref{fig:ablation_main}. Applying shuffling helps  maintaining global style consistency (Fig.~\ref{fig:ablation_main} (a)). Additionally, using DDIM inversion contributes to preserving the structure similar to the original image (Fig.~\ref{fig:ablation_main} (b)). Furthermore, our approach proves to be adaptable to different controls, such as lineart (Fig.~\ref{fig:ablation_main} (c)) and softedge (Fig.~\ref{fig:ablation_main} (d)) compared to depth used in RAVE (Fig.~\ref{fig:ablation_main} (e)). Even though there are style differences, these adjustments do not compromise the overall consistency.

%% file: sec_arxiv/6_discussion.tex
\subsection{Limitations}
\paragraph{Extreme shape editing in long videos} Shape editing is a challenging task in the field of video editing, with most existing methods struggling to maintain consistent shape transformations. Often, approaches that focus on shape editing rely on complex procedures like atlas editing \cite{lee2023shape}, and even these can lead to unsatisfactory outcomes. Our approach, in contrast to many current text-guided video editing models, is capable of handling shape edits from simple to extreme examples. For instance, our method can transform a wolf into a cat, bear, or dinosaur (as shown in Fig.~\ref{fig:teaser}), or can convert a boat into a jeep or a monkey into a bear (illustrated in Fig.~\ref{fig:main_results} of the main paper). Moreover, our method can handle \textbf{extreme} shape edits, such as transforming a car into a fire-truck, train, tractor, and so on.  However, while our method can handle these edits successfully,  it encounters limitations when performing \textbf{extreme} shape edits  as the video length increases. In particular, the ability of our method to maintain the distinct shape of these extreme objects weakens, resulting in some flickering. It is noteworthy that in cases of extreme editing, such as with the car-turn example, our method effectively manages shape transformations for up to 27 frames, beyond which the quality of the edit starts to degrade. This 27-frame threshold is significant as it represents the upper limit of the editing capabilities of many competing methods, such as FLATTEN \cite{cong2023flatten} (on RTX4090), for similar tasks.

\paragraph{Fine details flickering} 
Certain extreme shape editings (e.g., transforming the wolf into 'a unicorn') require high-frequency edits in the video (such as long and rich hair details of the unicorn). In such cases, flickering may occur as our model does not explicitly utilize pixel-level methods to address video deflickering. Furthermore, the unavoidable losses incurred during the compression in the encoding/decoding steps of latent diffusion models and the selection of inversion methods (DDIM inversion in our case) impact the quality of reconstructing fine details. Note that this is a common challenge present in existing approaches as well.

\section{Discussion and Conclusion}
\label{sec:conclusion}

In this paper, we present \texttt{RAVE}, a lightweight, fast method for zero-shot text-guided video editing using T2I diffusion models, leveraging spatio-temporal interaction with reduced computational costs. 
Through extensive comparisons, \texttt{RAVE} outperforms previous baselines in terms of temporal consistency, speed, and alignment with textual editing, for video editing including longer ones. \texttt{RAVE} is adaptable to various pre-trained models (\eg inpainting diffusion model, etc.), providing customizable video editing capabilities. We also highlight its potential for applications beyond video editing, such as consistent avatar generation or 3D texture editing, as part of our future work. By releasing our code and dataset for reproducibility, we encourage researchers to standardize evaluation.

\paragraph{Acknowledgements} This work was supported by NIH R01HD104624-01A1 and a gift from Meta.

%% file: sec_arxiv/supp.tex
\clearpage
\makeatletter
\renewcommand \thesection{S.\@arabic\c@section}
\renewcommand\thetable{S.\@arabic\c@table}
\renewcommand \thefigure{S.\@arabic\c@figure}
\makeatother
\setcounter{section}{0}
\setcounter{page}{1}
\maketitlesupplementary

\section{Existing Datasets in Video Editing Literature}
\label{sec:dataset_details}
 TokenFlow \cite{geyer2023tokenflow} utilized 61 text-video pairs sourced from DAVIS \cite{perazzi2016benchmark} and Internet Videos. Rerender \cite{yang2023rerender} employed test videos from Pexels \cite{pexels} and Pixabay \cite{pixabay}, while Text2Video-Zero \cite{khachatryan2023text2video} randomly selected 25 videos generated by CogVideo \cite{hong2023cogvideo}. 
 FateZero~\cite{qi2023fatezero} utilized videos from DAVIS and other in-the-wild videos, with text prompts created by the authors. Pix2Video~\cite{ceylan2023pix2video} employed a subset of videos from DAVIS, along with prompts acquired from previous works, some of which were generated by users. FLATTEN~\cite{cong2023flatten} used 16 videos from DAVIS and 37 videos from Videvo~\footnote{\href{https://www.videvo.net/}{Videvo} Last accessed: 2023-11-16.}, each with 4 prompts and 32 frames per video. Tune-A-Video~\cite{wu2023tune} employed 42 videos from DAVIS with 140 manually crafted text prompts. While some of these datasets are publicly available and others are not, there remains a notable lack of a standardized video dataset in the literature that includes a wide range of motions and longer-duration videos.

\section{Grid Trick}

\begin{figure}[h!]
    \centering
\includegraphics[width=1.0\linewidth]{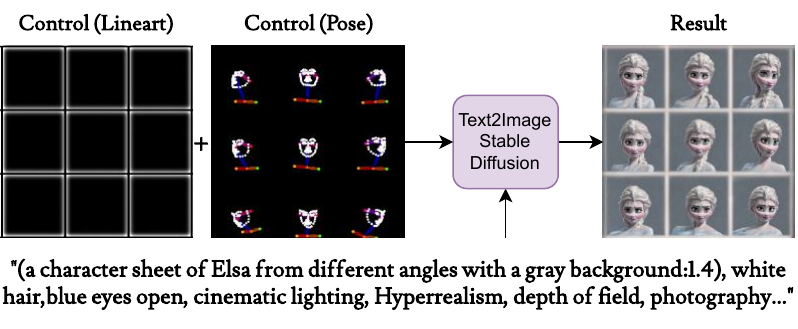}
    \caption{\textit{\textbf{Grid trick.}} 
} 
    \label{fig:grid_example}
\end{figure}
Often referred as the \textit{character sheet}, specifying desired characteristics in the text prompt and utilizing a grid (on the left) as a condition when employing ControlNet. The resulting output maintains the grid format during the editing process, ensuring consistent styles.

\section{User Study Details}
We conducted a user study involving 130 anonymous participants recruited from Prolific, a crowd-sourcing platform commonly utilized in research studies. The study focused on 23 randomly selected video-text pairs from our dataset. The comparison in the user study was made among Text2Video-Zero~\cite{khachatryan2023text2video}, Rerender~\cite{yang2023rerender}, Tokenflow~\cite{geyer2023tokenflow}, and our approach. Note that we employed Stable Diffusion v1.5 in all comparative analyses, including user study.

\begin{figure}[t!]
    \centering
\includegraphics[width=0.8\linewidth]{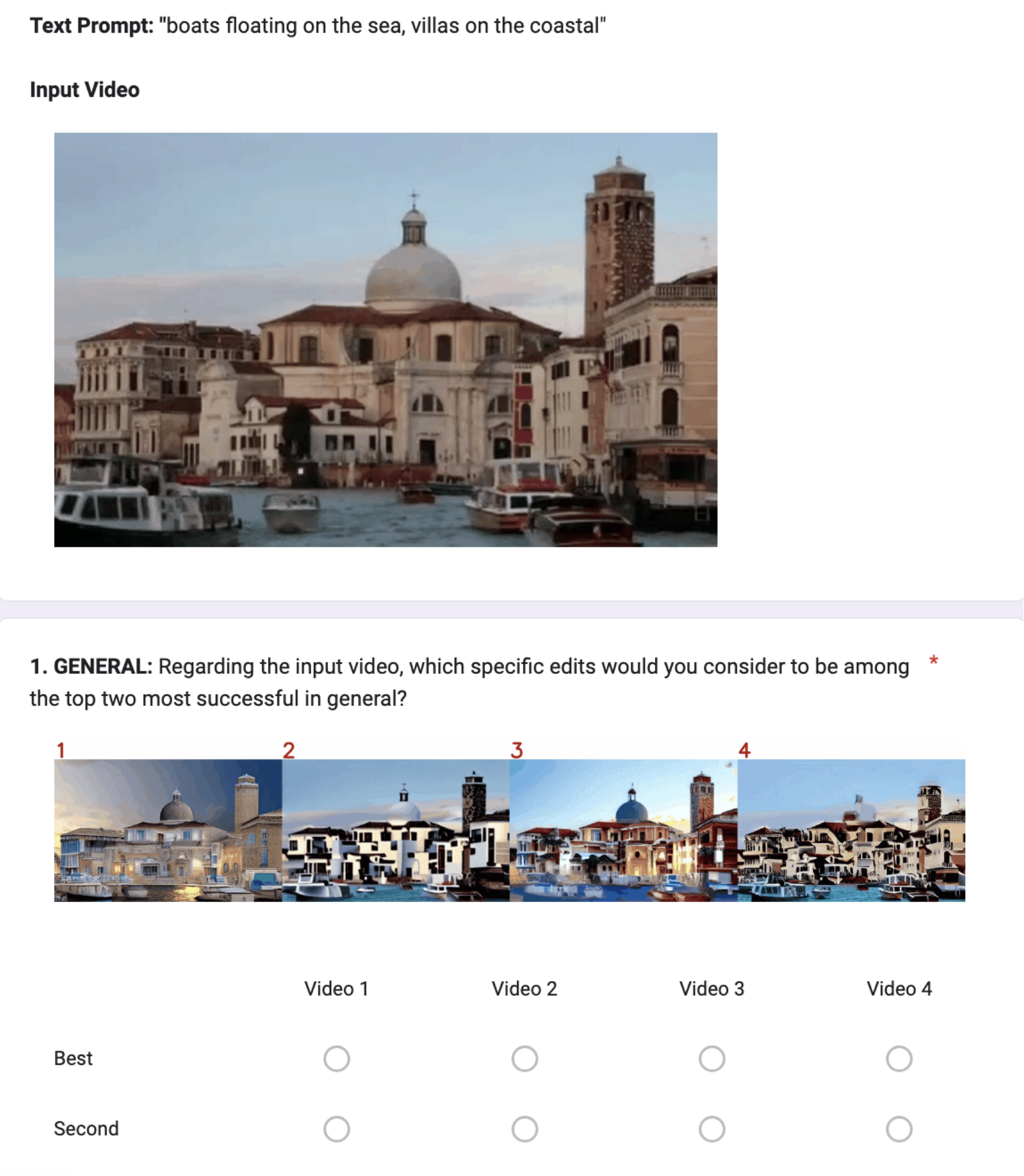}
    \caption{\textbf{\textit{A screenshot of the user study.}} }
\label{fig:survey}
\end{figure}

\begin{figure*}
    \centering
\includegraphics[width=0.8\linewidth]{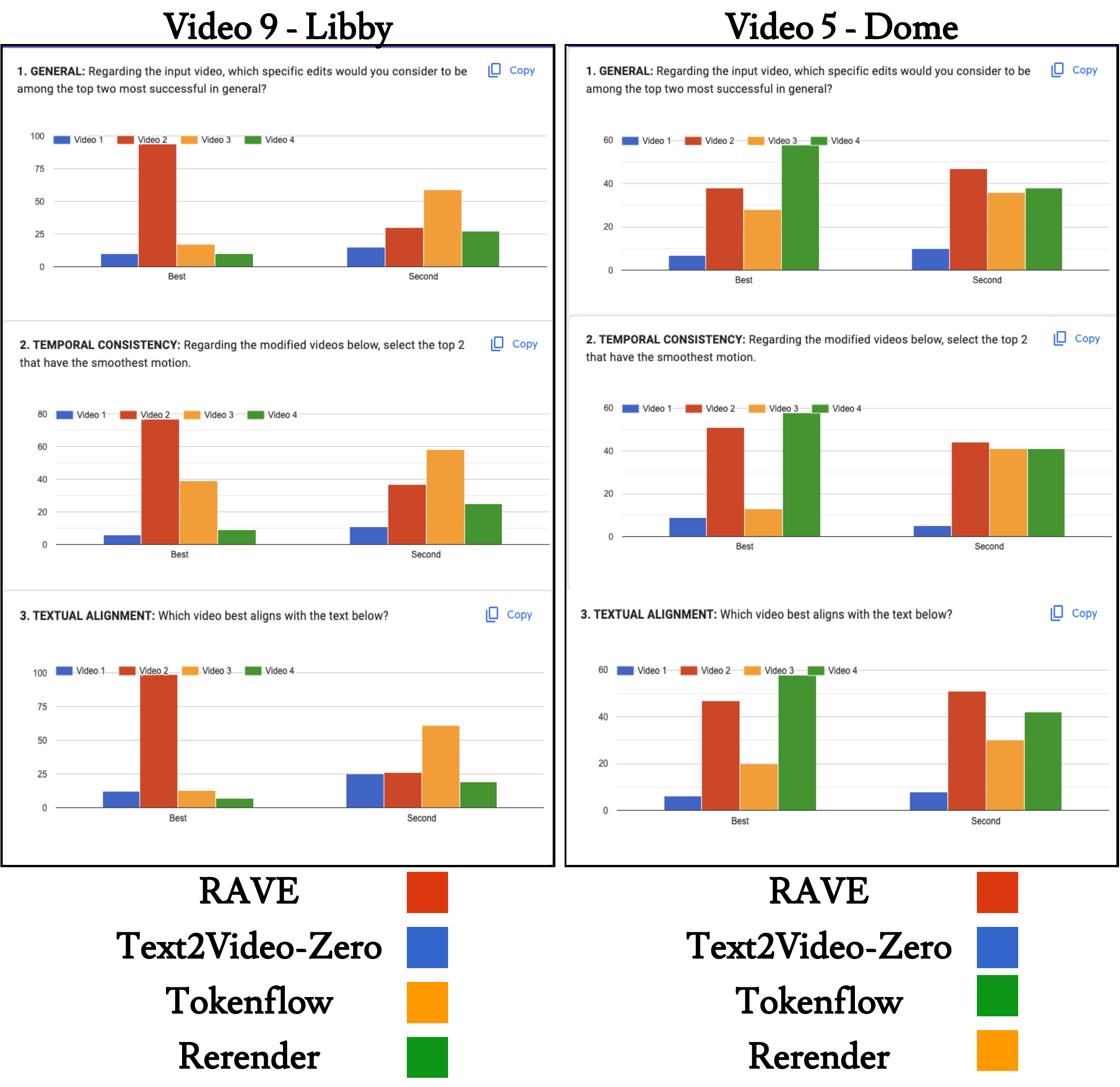}
    \caption{\textbf{\textit{Example results from user study.}} The results from two scenarios in our user study with 130 participants are illustrated. On the left, RAVE achieves the top-1 ranking on `libby' video, while on the right, Tokenflow secures the top-1 position. Each participant responded to three questions for each video.}
\label{fig:user_results}
\end{figure*}

Note that Tokenflow~\cite{geyer2023tokenflow} did not conduct a user study, whereas Rerender~\cite{yang2023rerender} surveyed 30 people to assess general editing capabilities. FateZero~\cite{qi2023fatezero} conducted a study with 20 participants, focusing on questions related to textual alignment, temporal consistency, and general editing capabilities. Additionally, Pix2Video~\cite{ceylan2023pix2video} conducted a survey with 37 participants, asking questions about general editing capabilities. FLATTEN~\cite{cong2023flatten} conducted a study with 16 participants, examining semantical, motional, temporal, and structural consistencies. In contrast, Tune-A-Video~\cite{wu2023tune} did not specify the number of users in their survey, concentrating on questions concerning temporal consistency and textual alignment. As opposed to previous works, our survey includes 130 anonymous participants.

\subsection{Questions}
We presented three questions to the participants, requesting them to rank the top two video edits among the four provided videos based on the following questions:
\begin{itemize}
    \item \textbf{Question 1 - General Editing (GE)}: ``Regarding the input video, which specific edits would you consider to be among the top two most successful in general?"
    \item \textbf{Question 2 - Temporal Consistency (TC)}: ``Regarding the modified videos below, select the top 2 that have the smoothest motion."
    \item \textbf{Question 3 - Textual Alignment (TA)}: ``Which video best aligns with the text below?"
\end{itemize}

The screenshot of the survey form for a single question and video is depicted in Fig.~\ref{fig:survey}. It's important to mention that for each user and video, the order of videos produced by each method is randomly shuffled to ensure an unbiased comparison.

\subsection{Further Analysis}

Note that we formulate a metric as the frequency of each method chosen among the top two edits, as provided in Table~\ref{tab:quant}. We provide the results of two examples (complete videos are available in the Supplementary Website) from our user study, one selected as the best and the other not selected, in response to Question 1 with that metric. (Fig.~\ref{fig:user_results})

We notice that videos with relatively stable backgrounds yield nearly evenly distributed selections among the different approaches. In contrast, videos featuring ego-exo motion and occlusions within dynamic scenes consistently demonstrate superior performance for our method compared to previous approaches. Please refer to the Supplementary Website for the complete videos.

\section{Further Results}

\subsection{Stable diffusion vs. realistic vision}
In our qualitative results, we utilize Realistic Vision V5.1 to harness its diverse editing capabilities (note that for comparison with other methods, we used Stable Diffusion v1.5 in order to have a fair comparison). As an ablation of using Realistic Vision model vs. Stable Diffusion, we also perform a comparison with the results obtained with each method. As observed, in both cases our method is able to apply the edits successfully, and temporal consistency is maintained. Please see Supplementary Website for examples. 

\subsection{Grid size vs metrics}

\begin{figure}[t!]
    \centering
    \includegraphics[width=1.0\linewidth]{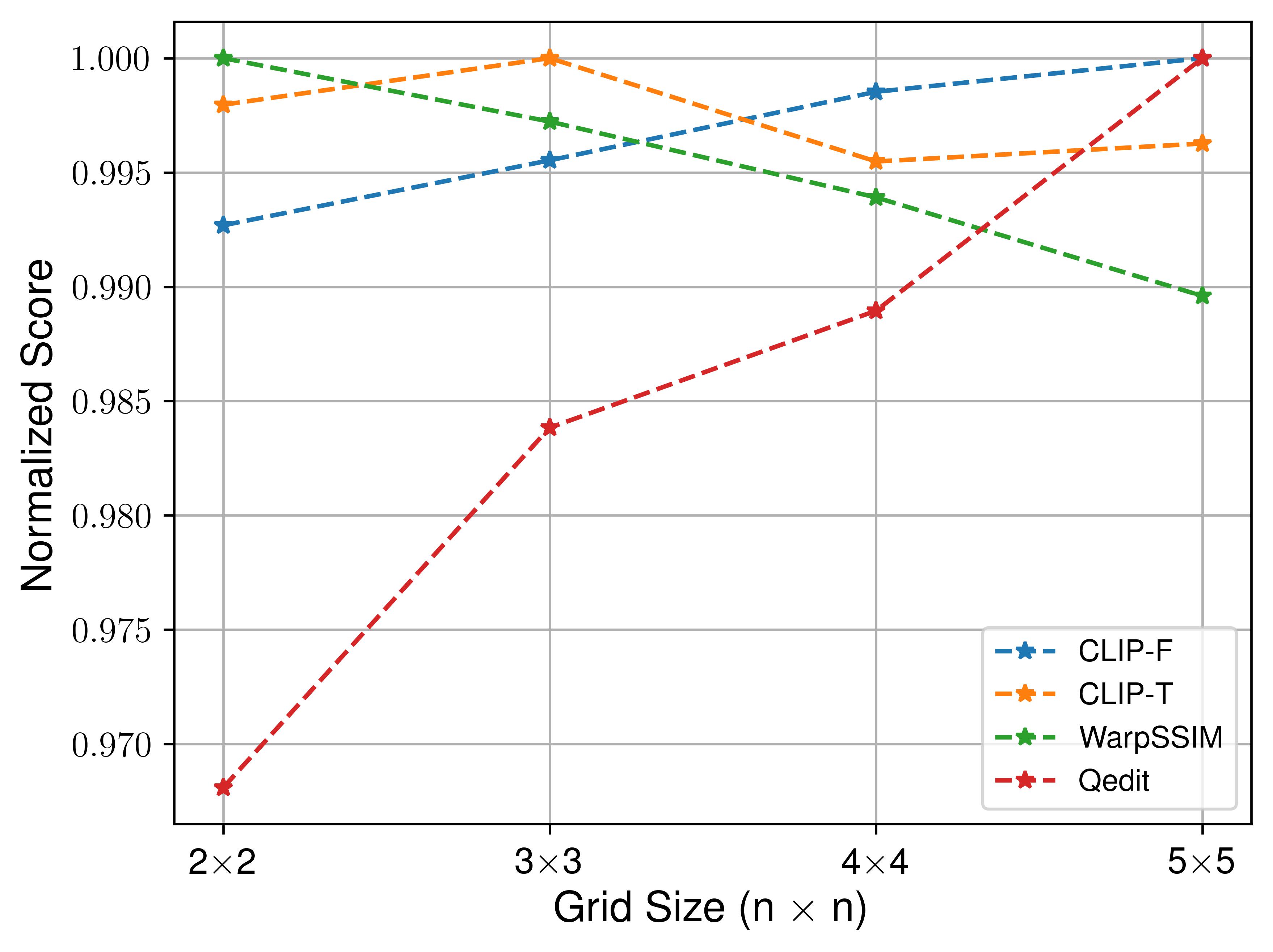}
    \caption{\textbf{\textit{Ablation study on grid size}}}
    \label{fig:ablation_gridsize}
\end{figure}

We conduct an additional ablation to explore the impact of grid size on the quality of editing. In Fig.~\ref{fig:ablation_gridsize}, the normalized metrics averaged over 90-frame videos are presented for grid sizes of $2\times 2$, $3\times 3$, $4\times 4$, and $5\times 5$. 
As anticipated, there is a noticeable improvement in CLIP-F and Q\textsubscript{edit} as the number of frames within a grid increases, attributed to the increased level of interaction that plays a more significant role in longer videos. Note that we chose $3\times 3$ in our experiments since the GPU requirement increases in proportion to the grid size. 

\section{Comparison with Keyframe Propogation}

\begin{figure}[h!]
    \centering
\includegraphics[width=1.0\linewidth]{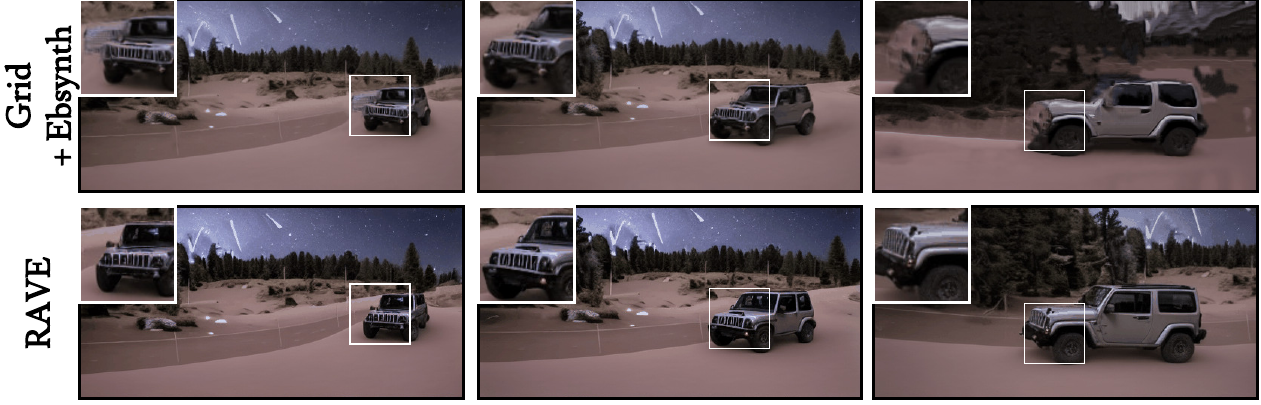}
\caption{\textbf{\textit{Comparison with Ebsynth.}}  We generated keyframes using the grid trick and employed Ebsynth to fill out between the keyframes.}
    \label{fig:comparison-ebysnth}
\end{figure}

We also evaluate our approach against a straightforward method. In this method, we produce keyframes using the grid trick and employ off-the-shelf tools such as Ebsynth~\cite{jamriska2018ebsynth} to fill in between these keyframes.

Figure~\ref{fig:comparison-ebysnth} illustrates an instance where Ebsynth is combined with the `Grid without shuffling' approach. It is evident that significant alterations occur in the structure of the car and the environment across consecutive frames of the keyframes in the Ebsynth method. In contrast, our approach demonstrates superior handling of temporal structural consistency.

%% file: main.bbl
\begin{thebibliography}{52}
\providecommand{\natexlab}[1]{#1}
\providecommand{\url}[1]{\texttt{#1}}
\expandafter\ifx\csname urlstyle\endcsname\relax
  \providecommand{\doi}[1]{doi: #1}\else
  \providecommand{\doi}{doi: \begingroup \urlstyle{rm}\Url}\fi

\bibitem[civ()]{civitai}
Civitai.
\newblock \url{https://civitai.com/}.
\newblock Accessed: 2023-11-16.

\bibitem[gri()]{gridtrick}
Gridtrick.
\newblock \url{https://web.archive.org/web/20231025170948/https://semicolon.dev/midjourney/how-to-make-consistent-characters}.
\newblock Archived: 2023-10-25.

\bibitem[pex()]{pexels}
Pexels.
\newblock \url{https://www.pexels.com/}.
\newblock Accessed: 2023-11-16.

\bibitem[pix()]{pixabay}
Pixabay.
\newblock \url{https://pixabay.com/}.
\newblock Accessed: 2023-11-16.

\bibitem[pro()]{prolific}
Prolific.
\newblock \url{https://www.prolific.com/}.
\newblock Accessed: 2023-11-18.

\bibitem[Avrahami et~al.(2022)Avrahami, Lischinski, and Fried]{avrahami2022blended}
Omri Avrahami, Dani Lischinski, and Ohad Fried.
\newblock Blended diffusion for text-driven editing of natural images.
\newblock In \emph{Proceedings of the IEEE/CVF Conference on Computer Vision and Pattern Recognition}, pages 18208--18218, 2022.

\bibitem[Avrahami et~al.(2023)Avrahami, Fried, and Lischinski]{avrahami2023blended}
Omri Avrahami, Ohad Fried, and Dani Lischinski.
\newblock Blended latent diffusion.
\newblock \emph{ACM Transactions on Graphics (TOG)}, 42\penalty0 (4):\penalty0 1--11, 2023.

\bibitem[Bain et~al.(2021)Bain, Nagrani, Varol, and Zisserman]{bain2021frozen}
Max Bain, Arsha Nagrani, G{\"u}l Varol, and Andrew Zisserman.
\newblock Frozen in time: A joint video and image encoder for end-to-end retrieval.
\newblock In \emph{ICCV}, pages 1728--1738, 2021.

\bibitem[Bar-Tal et~al.(2022)Bar-Tal, Ofri-Amar, Fridman, Kasten, and Dekel]{bar2022text2live}
Omer Bar-Tal, Dolev Ofri-Amar, Rafail Fridman, Yoni Kasten, and Tali Dekel.
\newblock Text2live: Text-driven layered image and video editing.
\newblock In \emph{European conference on computer vision}, pages 707--723. Springer, 2022.

\bibitem[Ceylan et~al.(2023)Ceylan, Huang, and Mitra]{ceylan2023pix2video}
Duygu Ceylan, Chun-Hao~P Huang, and Niloy~J Mitra.
\newblock Pix2video: Video editing using image diffusion.
\newblock In \emph{Proceedings of the IEEE/CVF International Conference on Computer Vision}, pages 23206--23217, 2023.

\bibitem[Cong et~al.(2023)Cong, Xu, Simon, Chen, Ren, Xie, Perez-Rua, Rosenhahn, Xiang, and He]{cong2023flatten}
Yuren Cong, Mengmeng Xu, Christian Simon, Shoufa Chen, Jiawei Ren, Yanping Xie, Juan-Manuel Perez-Rua, Bodo Rosenhahn, Tao Xiang, and Sen He.
\newblock Flatten: optical flow-guided attention for consistent text-to-video editing.
\newblock \emph{arXiv preprint arXiv:2310.05922}, 2023.

\bibitem[Couairon et~al.(2023)Couairon, Verbeek, Schwenk, and Cord]{couairon2023diffedit}
Guillaume Couairon, Jakob Verbeek, Holger Schwenk, and Matthieu Cord.
\newblock Diffedit: Diffusion-based semantic image editing with mask guidance.
\newblock In \emph{The Eleventh International Conference on Learning Representations}, 2023.

\bibitem[Gal et~al.(2023)Gal, Alaluf, Atzmon, Patashnik, Bermano, Chechik, and Cohen-or]{gal2023an}
Rinon Gal, Yuval Alaluf, Yuval Atzmon, Or Patashnik, Amit~Haim Bermano, Gal Chechik, and Daniel Cohen-or.
\newblock An image is worth one word: Personalizing text-to-image generation using textual inversion.
\newblock In \emph{The Eleventh International Conference on Learning Representations}, 2023.

\bibitem[Geyer et~al.(2023)Geyer, Bar-Tal, Bagon, and Dekel]{geyer2023tokenflow}
Michal Geyer, Omer Bar-Tal, Shai Bagon, and Tali Dekel.
\newblock Tokenflow: Consistent diffusion features for consistent video editing.
\newblock \emph{arXiv preprint arXiv:2307.10373}, 2023.

\bibitem[Hertz et~al.(2023)Hertz, Mokady, Tenenbaum, Aberman, Pritch, and Cohen-or]{hertz2023prompttoprompt}
Amir Hertz, Ron Mokady, Jay Tenenbaum, Kfir Aberman, Yael Pritch, and Daniel Cohen-or.
\newblock Prompt-to-prompt image editing with cross-attention control.
\newblock In \emph{The Eleventh International Conference on Learning Representations}, 2023.

\bibitem[Ho et~al.(2020)Ho, Jain, and Abbeel]{ho2020denoising}
Jonathan Ho, Ajay Jain, and Pieter Abbeel.
\newblock Denoising diffusion probabilistic models.
\newblock \emph{Advances in neural information processing systems}, 33:\penalty0 6840--6851, 2020.

\bibitem[Ho et~al.(2022{\natexlab{a}})Ho, Chan, Saharia, Whang, Gao, Gritsenko, Kingma, Poole, Norouzi, Fleet, et~al.]{ho2022imagen}
Jonathan Ho, William Chan, Chitwan Saharia, Jay Whang, Ruiqi Gao, Alexey Gritsenko, Diederik~P Kingma, Ben Poole, Mohammad Norouzi, David~J Fleet, et~al.
\newblock Imagen video: High definition video generation with diffusion models.
\newblock \emph{arXiv preprint arXiv:2210.02303}, 2022{\natexlab{a}}.

\bibitem[Ho et~al.(2022{\natexlab{b}})Ho, Salimans, Gritsenko, Chan, Norouzi, and Fleet]{ho2022video}
Jonathan Ho, Tim Salimans, Alexey Gritsenko, William Chan, Mohammad Norouzi, and David~J Fleet.
\newblock Video diffusion models.
\newblock In \emph{Advances in Neural Information Processing Systems}, pages 8633--8646. Curran Associates, Inc., 2022{\natexlab{b}}.

\bibitem[Hong et~al.(2023{\natexlab{a}})Hong, Lee, Jang, and Kim]{hong2023improving}
Susung Hong, Gyuseong Lee, Wooseok Jang, and Seungryong Kim.
\newblock Improving sample quality of diffusion models using self-attention guidance.
\newblock In \emph{Proceedings of the IEEE/CVF International Conference on Computer Vision}, pages 7462--7471, 2023{\natexlab{a}}.

\bibitem[Hong et~al.(2023{\natexlab{b}})Hong, Ding, Zheng, Liu, and Tang]{hong2023cogvideo}
Wenyi Hong, Ming Ding, Wendi Zheng, Xinghan Liu, and Jie Tang.
\newblock Cogvideo: Large-scale pretraining for text-to-video generation via transformers.
\newblock In \emph{The Eleventh International Conference on Learning Representations}, 2023{\natexlab{b}}.

\bibitem[Jamriska(2018)]{jamriska2018ebsynth}
Ondrej Jamriska.
\newblock Ebsynth: Fast example-based image synthesis and style transfer, 2018.

\bibitem[Jamri{\v{s}}ka et~al.(2019)Jamri{\v{s}}ka, Sochorov{\'a}, Texler, Luk{\'a}{\v{c}}, Fi{\v{s}}er, Lu, Shechtman, and S{\`y}kora]{jamrivska2019stylizing}
Ond{\v{r}}ej Jamri{\v{s}}ka, {\v{S}}{\'a}rka Sochorov{\'a}, Ond{\v{r}}ej Texler, Michal Luk{\'a}{\v{c}}, Jakub Fi{\v{s}}er, Jingwan Lu, Eli Shechtman, and Daniel S{\`y}kora.
\newblock Stylizing video by example.
\newblock \emph{ACM Transactions on Graphics (TOG)}, 38\penalty0 (4):\penalty0 1--11, 2019.

\bibitem[Kasten et~al.(2021)Kasten, Ofri, Wang, and Dekel]{kasten2021layered}
Yoni Kasten, Dolev Ofri, Oliver Wang, and Tali Dekel.
\newblock Layered neural atlases for consistent video editing.
\newblock \emph{ACM Transactions on Graphics (TOG)}, 40\penalty0 (6):\penalty0 1--12, 2021.

\bibitem[Kawar et~al.(2023)Kawar, Zada, Lang, Tov, Chang, Dekel, Mosseri, and Irani]{kawar2023imagic}
Bahjat Kawar, Shiran Zada, Oran Lang, Omer Tov, Huiwen Chang, Tali Dekel, Inbar Mosseri, and Michal Irani.
\newblock Imagic: Text-based real image editing with diffusion models.
\newblock In \emph{Proceedings of the IEEE/CVF Conference on Computer Vision and Pattern Recognition}, pages 6007--6017, 2023.

\bibitem[Khachatryan et~al.(2023)Khachatryan, Movsisyan, Tadevosyan, Henschel, Wang, Navasardyan, and Shi]{khachatryan2023text2video}
Levon Khachatryan, Andranik Movsisyan, Vahram Tadevosyan, Roberto Henschel, Zhangyang Wang, Shant Navasardyan, and Humphrey Shi.
\newblock Text2video-zero: Text-to-image diffusion models are zero-shot video generators.
\newblock In \emph{Proceedings of the IEEE/CVF International Conference on Computer Vision (ICCV)}, pages 15954--15964, 2023.

\bibitem[Lee et~al.(2023)Lee, Jang, Chen, Qiu, and Huang]{lee2023shape}
Yao-Chih Lee, Ji-Ze~Genevieve Jang, Yi-Ting Chen, Elizabeth Qiu, and Jia-Bin Huang.
\newblock Shape-aware text-driven layered video editing.
\newblock In \emph{Proceedings of the IEEE/CVF Conference on Computer Vision and Pattern Recognition}, pages 14317--14326, 2023.

\bibitem[Liew et~al.(2023)Liew, Yan, Zhang, Xu, and Feng]{liew2023magicedit}
Jun~Hao Liew, Hanshu Yan, Jianfeng Zhang, Zhongcong Xu, and Jiashi Feng.
\newblock Magicedit: High-fidelity and temporally coherent video editing.
\newblock \emph{arXiv preprint arXiv:2308.14749}, 2023.

\bibitem[Liu et~al.(2023)Liu, Zhang, Li, Lin, and Jia]{liu2023video}
Shaoteng Liu, Yuechen Zhang, Wenbo Li, Zhe Lin, and Jiaya Jia.
\newblock Video-p2p: Video editing with cross-attention control.
\newblock \emph{arXiv preprint arXiv:2303.04761}, 2023.

\bibitem[Meng et~al.(2021)Meng, He, Song, Song, Wu, Zhu, and Ermon]{meng2021sdedit}
Chenlin Meng, Yutong He, Yang Song, Jiaming Song, Jiajun Wu, Jun-Yan Zhu, and Stefano Ermon.
\newblock Sdedit: Guided image synthesis and editing with stochastic differential equations.
\newblock In \emph{ICLR}, 2021.

\bibitem[Mokady et~al.(2023)Mokady, Hertz, Aberman, Pritch, and Cohen-Or]{mokady2023null}
Ron Mokady, Amir Hertz, Kfir Aberman, Yael Pritch, and Daniel Cohen-Or.
\newblock Null-text inversion for editing real images using guided diffusion models.
\newblock In \emph{Proceedings of the IEEE/CVF Conference on Computer Vision and Pattern Recognition}, pages 6038--6047, 2023.

\bibitem[Molad et~al.(2023)Molad, Horwitz, Valevski, Acha, Matias, Pritch, Leviathan, and Hoshen]{molad2023dreamix}
Eyal Molad, Eliahu Horwitz, Dani Valevski, Alex~Rav Acha, Yossi Matias, Yael Pritch, Yaniv Leviathan, and Yedid Hoshen.
\newblock Dreamix: Video diffusion models are general video editors.
\newblock \emph{arXiv preprint arXiv:2302.01329}, 2023.

\bibitem[Parmar et~al.(2023)Parmar, Kumar~Singh, Zhang, Li, Lu, and Zhu]{parmar2023zero}
Gaurav Parmar, Krishna Kumar~Singh, Richard Zhang, Yijun Li, Jingwan Lu, and Jun-Yan Zhu.
\newblock Zero-shot image-to-image translation.
\newblock In \emph{ACM SIGGRAPH 2023 Conference Proceedings}, pages 1--11, 2023.

\bibitem[Perazzi et~al.(2016)Perazzi, Pont-Tuset, McWilliams, Van~Gool, Gross, and Sorkine-Hornung]{perazzi2016benchmark}
Federico Perazzi, Jordi Pont-Tuset, Brian McWilliams, Luc Van~Gool, Markus Gross, and Alexander Sorkine-Hornung.
\newblock A benchmark dataset and evaluation methodology for video object segmentation.
\newblock In \emph{Proceedings of the IEEE conference on computer vision and pattern recognition}, pages 724--732, 2016.

\bibitem[QI et~al.(2023)QI, Cun, Zhang, Lei, Wang, Shan, and Chen]{qi2023fatezero}
Chenyang QI, Xiaodong Cun, Yong Zhang, Chenyang Lei, Xintao Wang, Ying Shan, and Qifeng Chen.
\newblock Fatezero: Fusing attentions for zero-shot text-based video editing.
\newblock In \emph{Proceedings of the IEEE/CVF International Conference on Computer Vision (ICCV)}, pages 15932--15942, 2023.

\bibitem[Radford et~al.(2021)Radford, Kim, Hallacy, Ramesh, Goh, Agarwal, Sastry, Askell, Mishkin, Clark, et~al.]{radford2021learning}
Alec Radford, Jong~Wook Kim, Chris Hallacy, Aditya Ramesh, Gabriel Goh, Sandhini Agarwal, Girish Sastry, Amanda Askell, Pamela Mishkin, Jack Clark, et~al.
\newblock Learning transferable visual models from natural language supervision.
\newblock In \emph{International conference on machine learning}, pages 8748--8763. PMLR, 2021.

\bibitem[Ramesh et~al.(2022)Ramesh, Dhariwal, Nichol, Chu, and Chen]{ramesh2022hierarchical}
Aditya Ramesh, Prafulla Dhariwal, Alex Nichol, Casey Chu, and Mark Chen.
\newblock Hierarchical text-conditional image generation with clip latents.
\newblock \emph{arXiv preprint arXiv:2204.06125}, 2022.

\bibitem[Rombach et~al.(2022)Rombach, Blattmann, Lorenz, Esser, and Ommer]{rombach2022high}
Robin Rombach, Andreas Blattmann, Dominik Lorenz, Patrick Esser, and Bj{\"o}rn Ommer.
\newblock High-resolution image synthesis with latent diffusion models.
\newblock In \emph{Proceedings of the IEEE/CVF conference on computer vision and pattern recognition}, pages 10684--10695, 2022.

\bibitem[Ronneberger et~al.(2015)Ronneberger, Fischer, and Brox]{ronneberger2015u}
Olaf Ronneberger, Philipp Fischer, and Thomas Brox.
\newblock U-net: Convolutional networks for biomedical image segmentation.
\newblock In \emph{Medical Image Computing and Computer-Assisted Intervention--MICCAI 2015: 18th International Conference, Munich, Germany, October 5-9, 2015, Proceedings, Part III 18}, pages 234--241. Springer, 2015.

\bibitem[Ruiz et~al.(2023)Ruiz, Li, Jampani, Pritch, Rubinstein, and Aberman]{ruiz2023dreambooth}
Nataniel Ruiz, Yuanzhen Li, Varun Jampani, Yael Pritch, Michael Rubinstein, and Kfir Aberman.
\newblock Dreambooth: Fine tuning text-to-image diffusion models for subject-driven generation.
\newblock In \emph{Proceedings of the IEEE/CVF Conference on Computer Vision and Pattern Recognition}, pages 22500--22510, 2023.

\bibitem[Saharia et~al.(2022)Saharia, Chan, Saxena, Li, Whang, Denton, Ghasemipour, Gontijo~Lopes, Karagol~Ayan, Salimans, et~al.]{saharia2022photorealistic}
Chitwan Saharia, William Chan, Saurabh Saxena, Lala Li, Jay Whang, Emily~L Denton, Kamyar Ghasemipour, Raphael Gontijo~Lopes, Burcu Karagol~Ayan, Tim Salimans, et~al.
\newblock Photorealistic text-to-image diffusion models with deep language understanding.
\newblock \emph{Advances in Neural Information Processing Systems}, 35:\penalty0 36479--36494, 2022.

\bibitem[Shin et~al.(2023)Shin, Kim, Lee, Lee, and Yoon]{shin2023edit}
Chaehun Shin, Heeseung Kim, Che~Hyun Lee, Sang-gil Lee, and Sungroh Yoon.
\newblock Edit-a-video: Single video editing with object-aware consistency.
\newblock \emph{arXiv preprint arXiv:2303.07945}, 2023.

\bibitem[Singer et~al.(2023)Singer, Polyak, Hayes, Yin, An, Zhang, Hu, Yang, Ashual, Gafni, Parikh, Gupta, and Taigman]{singer2022make}
Uriel Singer, Adam Polyak, Thomas Hayes, Xi Yin, Jie An, Songyang Zhang, Qiyuan Hu, Harry Yang, Oron Ashual, Oran Gafni, Devi Parikh, Sonal Gupta, and Yaniv Taigman.
\newblock Make-a-video: Text-to-video generation without text-video data.
\newblock In \emph{The Eleventh International Conference on Learning Representations}, 2023.

\bibitem[Sohl-Dickstein et~al.(2015)Sohl-Dickstein, Weiss, Maheswaranathan, and Ganguli]{sohl2015deep}
Jascha Sohl-Dickstein, Eric Weiss, Niru Maheswaranathan, and Surya Ganguli.
\newblock Deep unsupervised learning using nonequilibrium thermodynamics.
\newblock In \emph{International conference on machine learning}, pages 2256--2265. PMLR, 2015.

\bibitem[Song et~al.(2021)Song, Meng, and Ermon]{song2021denoising}
Jiaming Song, Chenlin Meng, and Stefano Ermon.
\newblock Denoising diffusion implicit models.
\newblock In \emph{International Conference on Learning Representations}, 2021.

\bibitem[Teed and Deng(2020)]{teed2020raft}
Zachary Teed and Jia Deng.
\newblock Raft: Recurrent all-pairs field transforms for optical flow.
\newblock In \emph{Computer Vision--ECCV 2020: 16th European Conference, Glasgow, UK, August 23--28, 2020, Proceedings, Part II 16}, pages 402--419. Springer, 2020.

\bibitem[Tumanyan et~al.(2023)Tumanyan, Geyer, Bagon, and Dekel]{tumanyan2023plug}
Narek Tumanyan, Michal Geyer, Shai Bagon, and Tali Dekel.
\newblock Plug-and-play diffusion features for text-driven image-to-image translation.
\newblock In \emph{Proceedings of the IEEE/CVF Conference on Computer Vision and Pattern Recognition}, pages 1921--1930, 2023.

\bibitem[Valevski et~al.(2023)Valevski, Kalman, Molad, Segalis, Matias, and Leviathan]{valevski2023unitune}
Dani Valevski, Matan Kalman, Eyal Molad, Eyal Segalis, Yossi Matias, and Yaniv Leviathan.
\newblock Unitune: Text-driven image editing by fine tuning a diffusion model on a single image.
\newblock \emph{ACM Transactions on Graphics (TOG)}, 42\penalty0 (4):\penalty0 1--10, 2023.

\bibitem[Wang et~al.(2023)Wang, Xie, Liu, Chen, Cao, Wang, and Shen]{wang2023zero}
Wen Wang, Kangyang Xie, Zide Liu, Hao Chen, Yue Cao, Xinlong Wang, and Chunhua Shen.
\newblock Zero-shot video editing using off-the-shelf image diffusion models.
\newblock \emph{arXiv preprint arXiv:2303.17599}, 2023.

\bibitem[Wang et~al.(2004)Wang, Bovik, Sheikh, and Simoncelli]{wang2004image}
Zhou Wang, Alan~C Bovik, Hamid~R Sheikh, and Eero~P Simoncelli.
\newblock Image quality assessment: from error visibility to structural similarity.
\newblock \emph{IEEE transactions on image processing}, 13\penalty0 (4):\penalty0 600--612, 2004.

\bibitem[Wu et~al.(2023)Wu, Ge, Wang, Lei, Gu, Shi, Hsu, Shan, Qie, and Shou]{wu2023tune}
Jay~Zhangjie Wu, Yixiao Ge, Xintao Wang, Stan~Weixian Lei, Yuchao Gu, Yufei Shi, Wynne Hsu, Ying Shan, Xiaohu Qie, and Mike~Zheng Shou.
\newblock Tune-a-video: One-shot tuning of image diffusion models for text-to-video generation.
\newblock In \emph{Proceedings of the IEEE/CVF International Conference on Computer Vision}, pages 7623--7633, 2023.

\bibitem[Yang et~al.(2023)Yang, Zhou, Liu, , and Loy]{yang2023rerender}
Shuai Yang, Yifan Zhou, Ziwei Liu, , and Chen~Change Loy.
\newblock Rerender a video: Zero-shot text-guided video-to-video translation.
\newblock In \emph{ACM SIGGRAPH Asia Conference Proceedings}, 2023.

\bibitem[Zhang et~al.(2023)Zhang, Rao, and Agrawala]{zhang2023adding}
Lvmin Zhang, Anyi Rao, and Maneesh Agrawala.
\newblock Adding conditional control to text-to-image diffusion models.
\newblock In \emph{Proceedings of the IEEE/CVF International Conference on Computer Vision}, pages 3836--3847, 2023.

\end{thebibliography}
